\documentclass[journal]{IEEEtran}
\usepackage{amsmath,amsfonts}
\usepackage{algorithmic}
\usepackage{algorithm}
\usepackage{array}
\usepackage[caption=false,font=normalsize,labelfont=sf,textfont=sf]{subfig}
\usepackage{textcomp}
\usepackage{stfloats}
\usepackage{url}
\usepackage{verbatim}
\usepackage{graphicx}
\usepackage{cite}
\usepackage{booktabs,multirow, makecell, arydshln}
\usepackage{tablefootnote}

\begin{document}

\title{HAT: Hybrid Attention Transformer for \\Image Restoration}

\author{Xiangyu Chen, Xintao Wang, Wenlong Zhang, Xiangtao Kong, \\Yu Qiao,~\IEEEmembership{Senior~Member,~IEEE,} Jiantao Zhou,~\IEEEmembership{Senior~Member,~IEEE}, Chao Dong

\IEEEcompsocitemizethanks{\IEEEcompsocthanksitem Xiangyu Chen and Jiantao Zhou are with State Key Laboratory of Internet of Things for Smart City, University of Macau.
\IEEEcompsocthanksitem Xiangyu Chen, Yu Qiao and Chao Dong are with Shenzhen Institutes of Advanced Technology, Chinese Academy of Sciences, Shenzhen, China.
\IEEEcompsocthanksitem Xiangyu Chen, Wenlong Zhang, Xiangtao Kong, Chao Dong and Yu Qiao are with Shanghai Artificial Intelligence Laboratory, Shanghai, China. 
\IEEEcompsocthanksitem Xintao Wang is with the ARC Lab, Tencent PCG, Shenzhen, China.}
\thanks{Corresponding Authors: Jiantao Zhou (jtzhou@um.edu.mo) and Chao Dong (chao.dong@siat.ac.cn).}
}

\markboth{IEEE TRANSACTIONS ON PATTERN ANALYSIS AND MACHINE INTELLIGENCE}%
{Shell \MakeLowercase{\textit{et al.}}: A Sample Article Using IEEEtran.cls for IEEE Journals}


\maketitle

\begin{abstract}
Transformer-based methods have shown impressive performance in image restoration tasks, such as image super-resolution and denoising. 
However, we find that these networks can only utilize a limited spatial range of input information through attribution analysis. 
This implies that the potential of Transformer is still not fully exploited in existing networks. In order to activate more input pixels for better restoration, we propose a new Hybrid Attention Transformer (HAT). 
It combines both channel attention and window-based self-attention schemes, thus making use of their complementary advantages.
Moreover, to better aggregate the cross-window information, we introduce an overlapping cross-attention module to enhance the interaction between neighboring window features. 
In the training stage, we additionally adopt a same-task pre-training strategy to further exploit the potential of the model for further improvement. 
Extensive experiments have demonstrated the effectiveness of the proposed modules. We further scale up the model to show that the performance of the SR task can be greatly improved. 
Besides, we extend HAT to more image restoration applications, including real-world image super-resolution, Gaussian image denoising and image compression artifacts reduction.
Experiments on benchmark and real-world datasets demonstrate that our HAT achieves state-of-the-art performance both quantitatively and qualitatively.
Codes and models are publicly available at https://github.com/XPixelGroup/HAT.
\end{abstract}


\section{Introduction}
\IEEEPARstart{I}{mage} restoration (IR) is a classic problem in computer vision. It aims to reconstruct a high-quality (HQ) image from a given low-quality (LQ) input. 
Classic IR tasks encompass image super-resolution, image denoising, compression artifacts reduction, and etc.
Image restoration plays an important role in computer vision and has widespread application in areas such as AI photography~\cite{fatima2020ai}, surveillance imaging~\cite{zou2011very}, medical imaging~\cite{shi2013cardiac}, and image generation~\cite{karras2018progressive}.
Since deep learning has been successfully applied to IR tasks~\cite{srcnn_eccv,arcnn,dncnn}, numerous methods based on the convolutional neural network (CNN) have been proposed~\cite{fsrcnn,ircnn,rcan,rnan,esrgan,drunet} and almost dominate this field in the past few years. 
Recently, due to the success in natural language processing, Transformer~\cite{transformer} has attracted increasing attention in the computer vision community. 
After making rapid progress on high-level vision tasks~\cite{vit,swin_t,pvt}, Transformer-based methods are also developed for low-level vision tasks~\cite{ipt,uformer,restormer,edt,swinir}. 
A sucessful example is SwinIR~\cite{swinir}, which obtains a breakthrough improvement on IR tasks.

\begin{figure}[!t]
\centering
\includegraphics[width=1\linewidth]{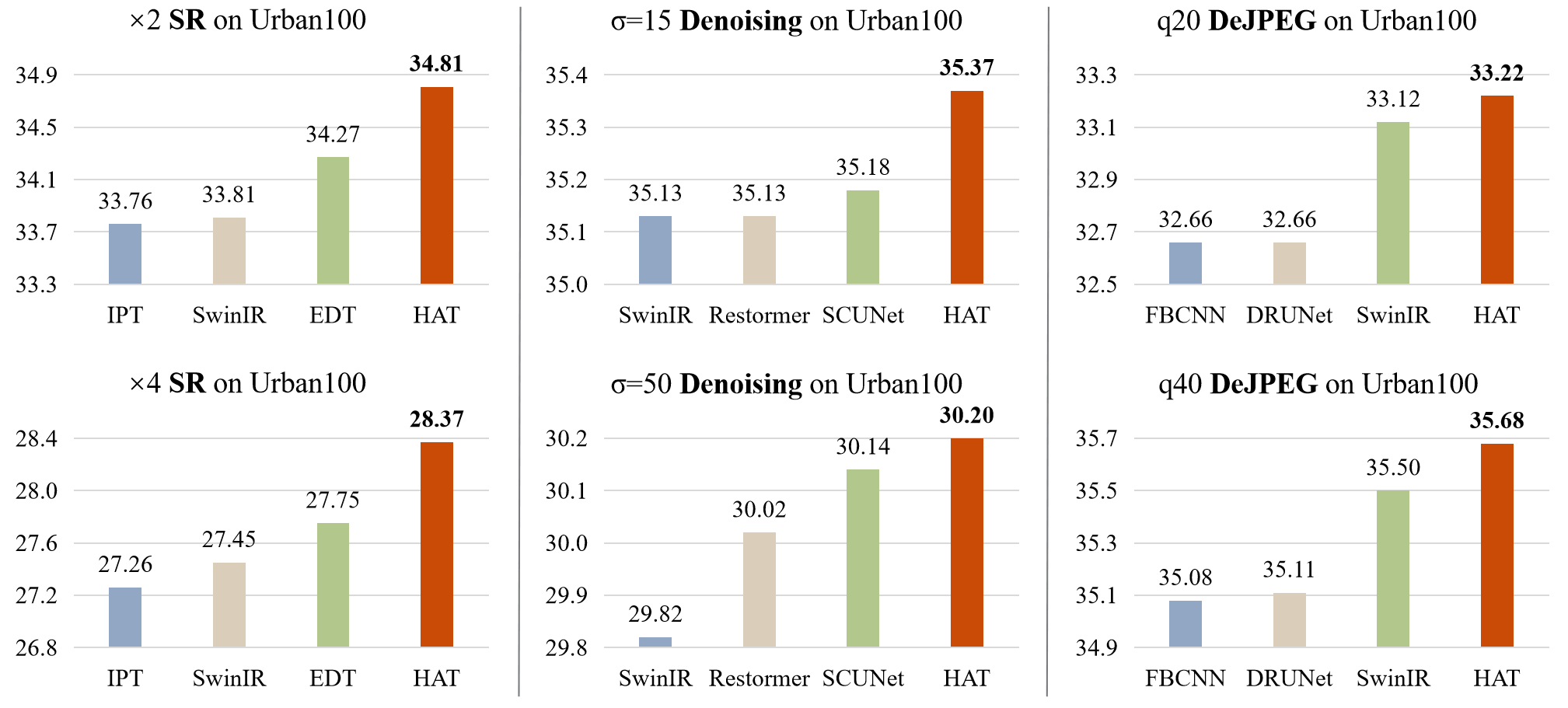}
\caption{Performance comparison of the proposed HAT on various image restoration tasks with the state-of-the-art methods.}
\label{performance}
\vspace{-0.5cm}
\end{figure}

Despite its success, existing work has rarely discussed why Transformer outperforms CNN.  
An intuitive explanation provided in prior study is that Transformer benefits from the self-attention mechanism, allowing it to leverage long-range information~\cite{swinir}.
To verify whether this is indeed the case for image restoration, we take image super-resolution (SR) as an example task and employ an attribution analysis method — Local Attribution Map (LAM)~\cite{lam} to examine the range of information used in SwinIR.
Interestingly, we find that although SwinIR achieves higher average quantitative performance, it does NOT utilize more input pixels than CNN-based methods (e.g., RCAN~\cite{rcan}), as shown in Fig.~\ref{lam}. 
This contradicts the conclusion in LAM~\cite{lam} that there is a positive correlation between the range of information a network uses and its reconstruction performance.
Since the aforementioned conclusion is primarily derived from networks of the same type (i.e., CNNs), we believe that the superior performance of SwinIR can be attributed to its stronger ability to model local information compared to CNN. 
However, it is also limited by the restricted range of information it utilizes, leading to inferior results on samples where broader contextual information could produce better outcomes. 
Additionally, we observe block artifacts in the intermediate features of SwinIR, as shown in Fig.~\ref{feature}, suggesting that the shifted window mechanism does not fully achieve cross-window information interaction. This may be one of the reasons why SwinIR does not achieve better long-range information utilization.

To address the above-mentioned limitations of the existing IR Transformer and further develop the potential of such networks, we propose a Hybrid Attention Transformer, namely HAT. It combines channel attention and self-attention schemes, in order to take advantage of the former's capability in using global information and the powerful representative ability of the latter.
Besides, we introduce an overlapping cross-attention module to achieve more direct interaction of adjacent window features. Benefiting from these designs, our model can activate more pixels for reconstruction and thus obtains significant performance improvement.
Since Transformers do not have an inductive bias like CNNs, large-scale data pre-training is important to unlock the potential of such models. In this paper, we provide an effective \textit{same-task pre-training} strategy. Different from IPT~\cite{ipt} using multiple restoration tasks for pre-training and EDT~\cite{edt} utilizing multiple degradation levels of a specific task for pre-training, we directly perform pre-training using large-scale dataset on the same task. 
We believe that large-scale data is what really matters for pre-training. Experimental results show the superiority of our strategy. 
Equipped with the above designs, HAT surpasses the state-of-the-art methods by a large margin on SR, as well as several other image restoration tasks, as shown in Fig.~\ref{performance}. 
Overall, our main contributions are four-fold: 
\begin{itemize}
\item We design a Hybrid Attention Transformer (HAT) that combines self-attention, channel attention and a new overlapping cross-attention for high-quality image restoration.

\item We propose an effective same-task pre-training strategy to further exploit the potential of SR Transformer and show the importance of large-scale data pre-training. 

\item Our method significantly outperforms existing state-of-the-art methods on the SR task. By further scaling up HAT to build a large model, we greatly extend the performance upper bound of the SR task. 

\item Our method also achieves state-of-the-art performance on image denoising and compression artifacts reduction, showing its superiority on various image restoration tasks. 
\end{itemize} 

A preliminary version of this work was presented at CVPR2023~\cite{hat_conf}. The present work expands upon the initial version in several significant ways. 
Firstly, we provide a theoretic illustration of LAM and augment the analysis with CEM results. This can facilitate readers’ understanding of the motivation behind our method and the rationality of its design. 
Secondly, we investigate a flexible plain architecture of HAT for application to various IR tasks, allowing us to further explore the potential breadth of this method. 
Thirdly, we extend HAT to real-world image super-resolution based on practical degradation models. The promising results show the potential of HAT for real-world applications. 
Additionally, we further extend HAT for image denoising and compression artifacts reduction. Extensive experiments show that our method achieves state-of-the-art performance on several IR tasks. 

\section{Related Work}
\subsection{Image Super-Resolution} 
Since SRCNN~\cite{srcnn_eccv} first introduces deep convolution neural networks (CNNs) to the image SR task and obtains superior performance over conventional SR methods, numerous deep networks~\cite{srcnn_tpami,fsrcnn,pixelshuffle,vdsr,edsr,rdn,rcan,san,han,nlsn,swinir,edt} have been proposed for SR to further improve the reconstruction quality. For instance, many methods apply more elaborate convolution module designs, such as residual block~\cite{srgan,edsr} and dense block~\cite{esrgan,rdn}, to enhance the model representation ability. Several works explore more different frameworks like recursive neural network~\cite{drcn,drrn} and graph neural network~\cite{ignn}. To improve perceptual quality, \cite{srgan,esrgan,ranksrgan,realesrgan,femasr} introduce adversarial learning to generate more realistic results. By using attention mechanism, \cite{rcan,san,rnan,nlrn,han,nlsn} achieve further improvement in terms of reconstruction fidelity. Recently, a series of Transformer-based networks~\cite{ipt,swinir,edt} are proposed and constantly refresh the state-of-the-art of SR task, showing the powerful representation ability of Transformer. 

To deepen the understanding of SR networks, several studies \cite{lam,ddr,faig,dropoutSR,srga,cem} are conducted to analyze and interpret their working mechanism. LAM~\cite{lam} adopts the integral gradient method to explore which input pixels contribute to the final performance. DDR~\cite{ddr} reveals the deep semantic representations in SR networks based on deep feature dimensionality reduction and visualization. FAIG~\cite{faig} is proposed to find discriminative filters for specific degradations in blind SR. \cite{dropoutSR} introduces channel saliency map to show that Dropout can help prevent co-adapting for real SR networks. SRGA~\cite{srga} aims to evaluate the generalization ability of SR methods. CEM~\cite{cem} interprets the low-level vision models based on  causal effect theory. In this work, we exploit LAM~\cite{lam} and CEM~\cite{cem} to analyze and understand the behavior of different networks. 

\subsection{Vision Transformer}
Recently, Transformer~\cite{transformer} has attracted the attention of computer vision community due to its success in the field of natural language processing. A series of Transformer-based methods~\cite{vit,localvit,cvt,pvt,swin_t,shufflet,cswin,twins,palet,ceit,uniformer,moa} have been developed for high-level vision tasks, including image classification~\cite{swin_t,vit,localvit,ramachandran2019stand,halonet}, object detection~\cite{liu2020deep,touvron2021training,swin_t,detr,twins}, segmentation~\cite{wu2020visual,pvt,dat,cao2022swin}, \textit{etc}. Although vision Transformer has shown its superiority on modeling long-range dependency~\cite{vit,cka}, there are still many works demonstrating that the convolution can help Transformer achieve better visual representation~\cite{cvt,vitc,hrformer,ceit,uniformer}. Due to the impressive performance, Transformer has also been introduced for low-level vision tasks~\cite{ipt,uformer,swinir,vsrt,maxim,restormer,vrt,edt}. Specifically, IPT~\cite{ipt} develops a ViT-style network and introduces multi-task pre-training for image processing. SwinIR~\cite{swinir} proposes an image restoration Transformer based on \cite{swin_t}. VRT\cite{vrt} introduces Transformer-based networks to video restoration. EDT\cite{edt} adopts self-attention mechanism and multi-related-task pre-training strategy to further refresh the state-of-the-art of SR. However, existing works still cannot fully exploit the potential of Transformer, while our method can activate more input pixels for better reconstruction.

\subsection{Deep Networks for Image Restoration}
Image restoration, which aims to recover high-quality images from degraded inputs, has seen significant progress with the rise of deep learning. Early successes are achieved in tasks like image super-resolution~\cite{srcnn_tpami}, image denoising~\cite{dncnn}, and compression artifact reduction~\cite{arcnn}. Numerous CNN-based networks have since been proposed for image restoration~\cite{ircnn,ffdnet,rdn_pami,Gopro,mprnet,nafnet,uformer,restormer,maxim,swinir,edt,tape,scunet}. Before the advent of Transformers in low-level vision tasks, CNNs dominated the field. For example, ARCNN~\cite{arcnn} employs stacked convolutional layers to address JPEG compression artifacts, and DnCNN~\cite{dncnn} combines convolution with batch normalization for image denoising. RDN~\cite{rdn_pami} introduces a residual dense CNN architecture, excelling in various restoration tasks. A comprehensive investigation of CNN-based methods for SR can be found in~\cite{journey}.
As Transformers have gained prominence in computer vision, Transformer-based image restoration methods have emerged. SwinIR~\cite{swinir}, built on the Swin Transformer~\cite{swin_t}, demonstrates excellent performance on image super-resolution, denoising, and JPEG artifact reduction. Uformer~\cite{uformer} introduces a U-Net-style Transformer for diverse restoration tasks, while Restormer~\cite{restormer} innovates with transposed self-attention to achieve state-of-the-art results. SCUNet~\cite{scunet} combines CNNs and Transformers to create a highly effective denoising network. Transformer-based networks have demonstrated superior performance compared to previous CNN-based methods. In this paper, we introduce a hybrid attention mechanism, further improving the performance of image restoration Transformer.

\section{Motivation}
\label{Motivation}
Swin Transformer~\cite{swin_t} has already demonstrated excellent performance in image restoration tasks~\cite{swinir}. We are thus eager to understand what makes it superior to CNN-based methods and what its potential shortcomings are that could be improved. To explore these questions, we seek to derive insights using interpretability tools and visualization analysis. Given that existing IR Transformer shows remarkable advancements on SR, and that the analytical tools developed for SR are more mature, we focus our analysis primarily on SR. In this section, we present the motivation behind our approach. We begin by reviewing the LAM method, an attribution analysis tool for SR networks. Next, we apply LAM to several classical SR networks and augment the results with causal analysis using the Causal Effect Map (CEM)~\cite{cem}. Finally, we present feature visualization results that provide additional insights.

 \subsection{An Overview of LAM}

Local Attribution Map (LAM)~\cite{lam} is an attribution analysis method tailored for image SR. It extends the classical integrated gradients~\cite{sundararajan2017axiomatic} by introducing a task-specific baseline and path function suited to the characteristics of SR networks, which focus on reconstructing high-frequency details such as textures and edges.

Formally, let $F: \mathbb{R}^n \rightarrow \mathbb{R}^k$ denote an SR network, $I$ be the input image, and $I'$ be the baseline input. The path-integrated gradients along the $i$-th dimension are computed as:
\begin{equation}
    \mathsf{PathIntegratedGrads}^{\lambda}_i(I) := \int_0^1 \frac{\partial F(\lambda(\theta))}{\partial \lambda_i(\theta)} \cdot \frac{\partial \lambda_i(\theta)}{\partial \theta} \, d\theta,
\end{equation}
where $\lambda(\theta)$ defines a smooth interpolation from $I'$ to $I$. Unlike classification tasks where $I'$ is typically a black image, LAM defines $I'$ as a blurred version of $I$:
\begin{equation}
    I' = \omega(\sigma) \otimes I,
\end{equation}
where $\omega(\sigma)$ is a Gaussian kernel with standard deviation $\sigma$, and $\otimes$ denotes convolution. The path function $\lambda_{\mathrm{pb}}(\theta)$ progressively reduces the blur:
\begin{equation}
    \lambda_{\mathrm{pb}}(\theta) = \omega(\sigma - \theta \sigma) \otimes I,
\end{equation}
so that $\lambda_{\mathrm{pb}}(0) = I'$ and $\lambda_{\mathrm{pb}}(1) = I$.

To assess how input pixels contribute to the reconstruction of high-frequency details, LAM applies a gradient-based detector $D_{xy}$ (e.g., Gabor filter~\cite{granlund1978search}) to a patch centered at $(x, y)$:
\begin{equation}
    D_{xy}(I) = \sum_{m,n \in \mathcal{P}_{xy}} \nabla_{mn} I,
\end{equation}
where $\nabla_{mn} I$ denotes the image gradient and $\mathcal{P}_{xy}$ is the local patch. The final attribution score for the $i$-th pixel is then:
\begin{equation}
    \mathsf{LAM}_{F,D}(\lambda_{\mathrm{pb}})_i := \int_0^1 \frac{\partial D(F(\lambda_{\mathrm{pb}}(\theta)))}{\partial \lambda_{\mathrm{pb}}(\theta)_i} \cdot \frac{\partial \lambda_{\mathrm{pb}}(\theta)_i}{\partial \theta} \, d\theta.
\end{equation}

As illustrated in Fig.~\ref{lam}, LAM provides spatial heatmaps that visualize the importance of each input pixel to the reconstruction of structural details. To quantify the spatial extent of utilized information, LAM further defines the \textit{Diffusion Index} (DI), a metric derived from the Gini coefficient~\cite{Gini}, where a larger DI reflects a broader and more uniform use of input pixels, often correlating with better reconstruction quality~\cite{lam}.

While LAM is effective for SR, its extension to other restoration tasks such as denoising or compression artifact reduction remains an open challenge. This is because LAM depends on a well-defined baseline and a continuous degradation path, both of which are difficult to establish in tasks with stochastic or non-differentiable degradations (e.g., Gaussian noise). Defining a “more noisy” baseline and constructing a semantically meaningful interpolation path is non-trivial and may lead to unstable or unreliable attribution results. We thus conduct LAM experiments merely on SR.

\subsection{Interpretability Analysis}

\begin{figure}[!t]
\centering
\includegraphics[width=1\linewidth]{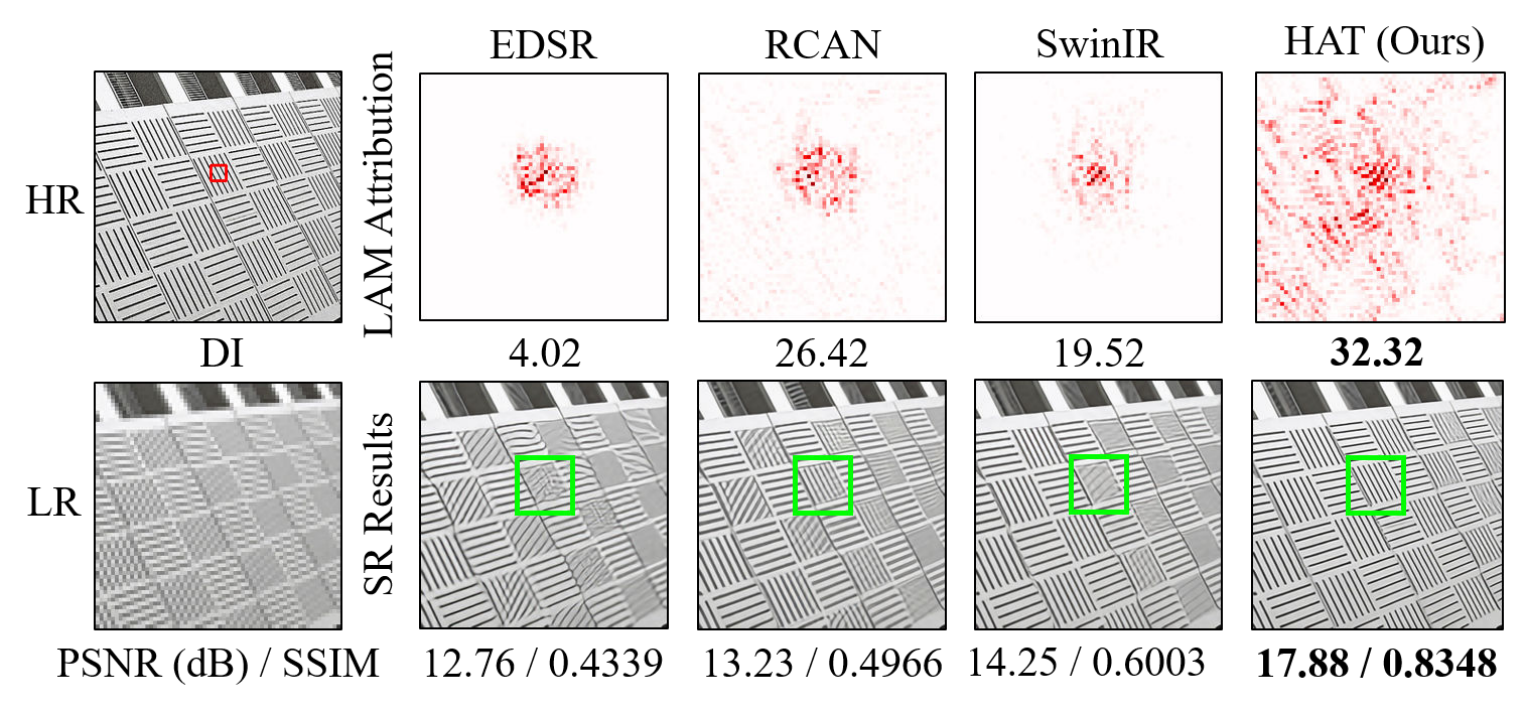}
\caption{LAM~\cite{lam} results of different networks. SwinIR utilizes less information compared to RCAN, while HAT uses the most pixels for reconstruction.}
\label{lam}
\end{figure}

We first employ LAM to perform attribution analysis on several classic SR networks, as shown in Fig.~\ref{lam}. Intuitively, SR networks that utilize more input information achieve superior reconstruction performance. This relationship is clearly observed in the comparison between EDSR~\cite{edsr} and RCAN~\cite{rcan}. However, this conclusion is the opposite in the comparison between RCAN and SwinIR. SwinIR achieves better reconstruction results despite utilizing significantly less input information. First, this LAM observation contradicts the intuition in existing literature~\cite{swinir}, which suggests that Transformers perform better by more effectively modeling long-range dependency. Second, it means that SwinIR, which employs a window-based self-attention mechanism, excels at capturing local information and can achieve superior performance with less input information. Additionally, we observe that SwinIR produces incorrect texture reconstruction in case where RCAN successfully restores the texture, which may be attributed to SwinIR’s limited information utilization.

\begin{figure}[!t]
\centering
\includegraphics[width=1\linewidth]{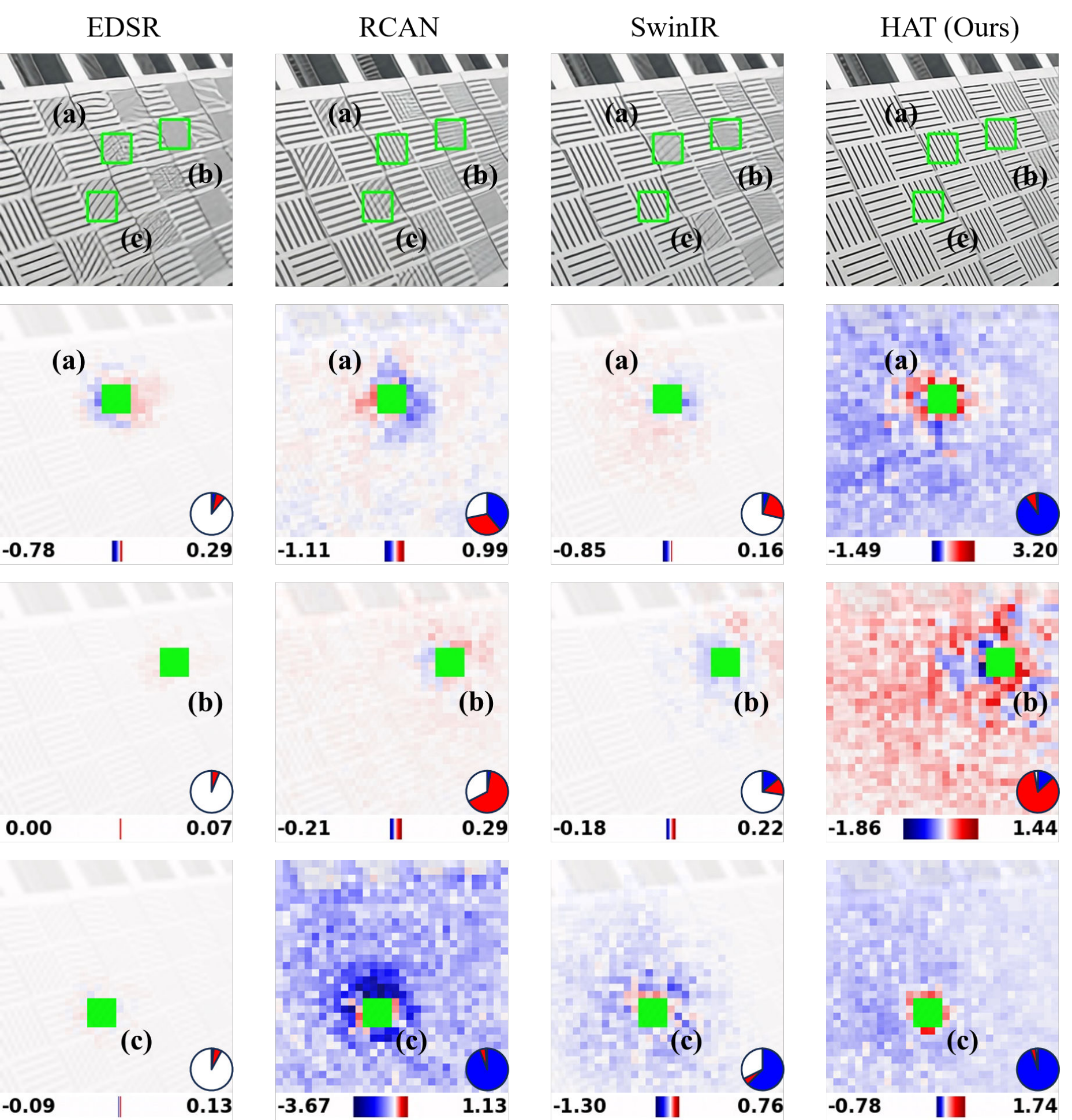}
\caption{CEM~\cite{cem} results of different networks. Activating more input information for Transformer is crucial to the reconstruction performance.}
\label{cem}
\end{figure}

To further analyze these behaviors, we use the Causal Effect Map (CEM)~\cite{cem}, which measures how each input patch affects the reconstruction of the region of interest (ROI), marked in green in Fig.~\ref{cem}. Patches with positive or negative causal effects are shown in red and blue, respectively. The pie chart indicates the distribution of effects, and the color bar reflects the magnitude.
Although all examples in Fig.~\ref{cem} share similar striped patterns, the amount of useful detail in the ROI of the LQ input varies. For reference, all examples are reconstructed from the same LQ image shown in the bottom-left of Fig.~\ref{lam}. In example (b), the ROI contains almost no visible texture in the input. The model must therefore rely on surrounding patches to infer the structure, resulting in many patches showing positive causal effects. In contrast, examples (a) and (c) contain clearer directional patterns in the ROI. In these cases, the model can mainly rely on local information, and using too much external context may introduce conflicting signals, leading to negative causal effects.
This behavior aligns with causal reasoning:  
when the internal evidence is weak, external information supports the reconstruction;  
when internal evidence is strong, external input may act as a confounder.
Our HAT adapts well to both situations. It activates a wide range of input pixels when needed (e.g., example b), and focuses more locally when the ROI already contains useful information (e.g., examples a and c). This flexibility results in more accurate textures and sharper edges across diverse scenarios.
In conclusion, we posit that the performance of SR networks depends not only on the quantity of activated information but also on how effectively the model adapts to the local content. Enhancing both the usable spatial range and context-awareness is essential for developing more powerful SR models.

\begin{figure}[!t]
\centering
\includegraphics[width=1\linewidth]{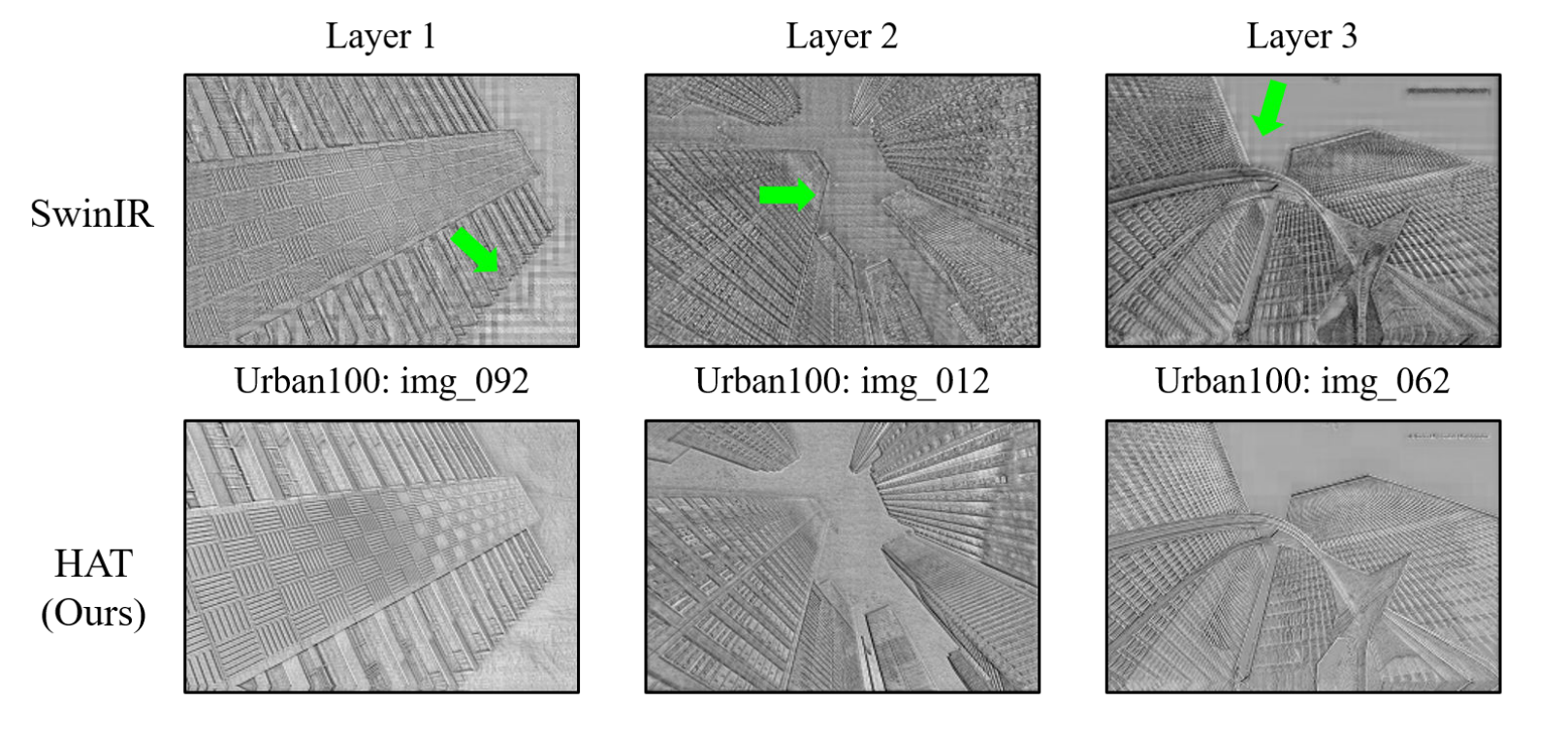}
\caption{
Intermediate features visualization. ``Layer N'' means the intermediate features after the $N_{th}$ layer (\textit{i.e.}, RSTB in SwinIR and RHAG in HAT.) 
}
\label{feature}
\vspace{-5pt}
\end{figure}

\subsection{Feature Visualization}
SwinIR, as a new architecture distinct from traditional CNN designs, motivates us to examine its intermediate features to gain further insights. 
As shown in Figure ~\ref{feature}, we observe noticeable block artifacts in SwinIR. Interestingly, the size of these blocks coincides with the window size, suggesting that these artifacts are likely caused by the window partitioning mechanism. 
This indicates that the shifted window approach may be insufficient for effectively integrating information across windows. 
This limitation could be one of the reasons why SwinIR fails to utilize more pixels for reconstruction, as evidenced in Figures ~\ref{lam} and ~\ref{cem}. 
Several studies on high-level vision tasks have also pointed out that enhancing connections between windows can improve window-based self-attention mechanisms~\cite{cswin,shufflet,palet,moa}. 
Consequently, we enhance the interaction of information across windows in our method. We can see that the block artifacts in the intermediate features of our HAT are significantly alleviated.


\begin{figure*}[!t]
\centering
\includegraphics[width=0.95\linewidth]{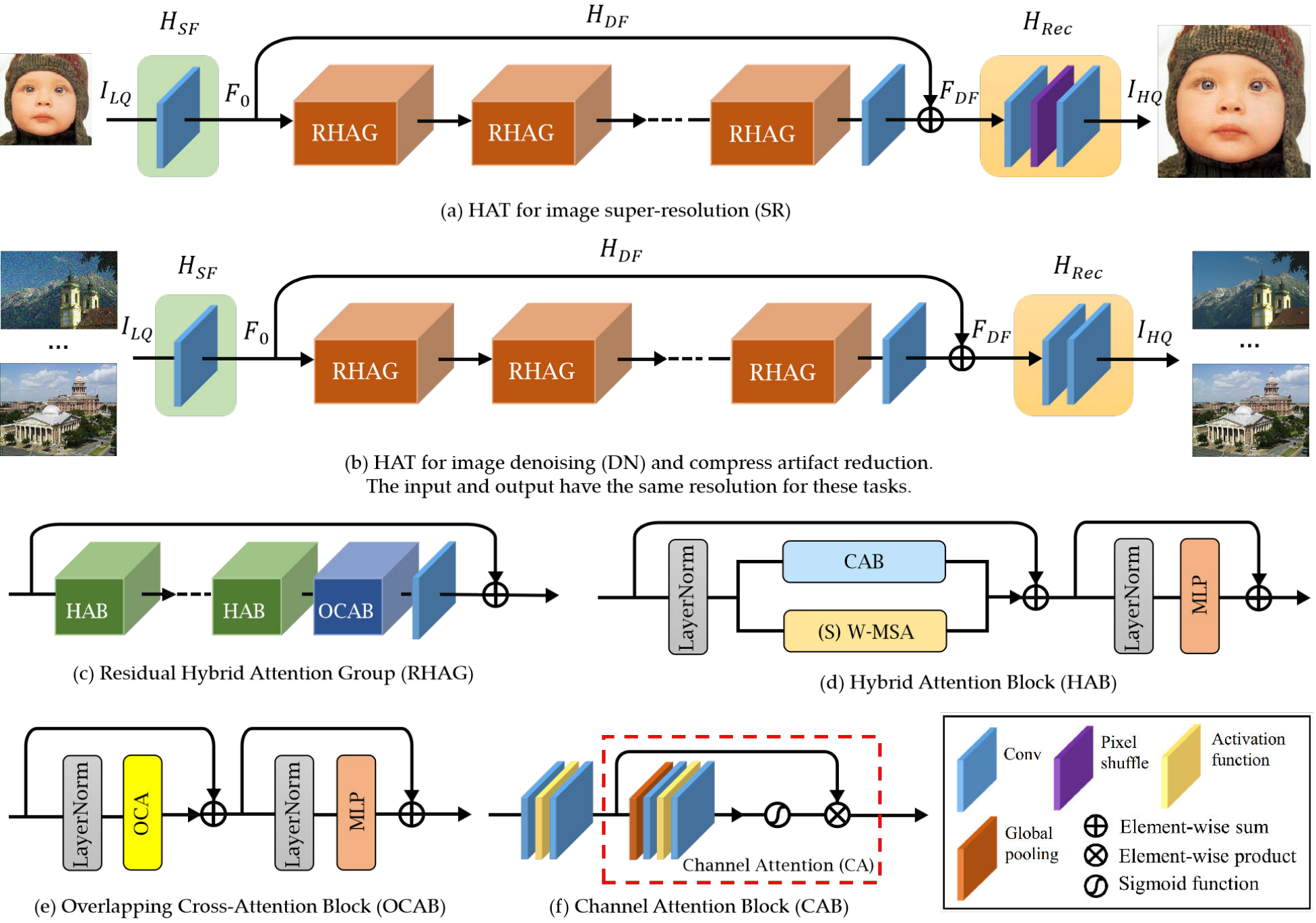}
\caption{The overall architecture of HAT and the structure of RHAG and HAB.}
\label{Network Structure}
\vspace{-8pt}
\end{figure*}

\section{Methodology}
Based on the above analysis, we aim to design a better image restoration network by enhancing the ability of the existing Transformer model to efficiently utilize more input information, integrating global information, and improving the cross-window interaction. In this section, we provide a detailed introduction to our approach, HAT, including the overall architecture, key module designs, training strategy, implementation details, as well as discussions with other methods.

\subsection{Network Structure of HAT}
The overall network structure of HAT follows the classic Residual in Residual (RIR) architecture similar to \cite{rcan,swinir}. 
As shown in Fig.~\ref{Network Structure}, HAT consists of three parts, including shallow feature extraction, deep feature extraction and image reconstruction.   
Concretely, for a given low-quality (LQ) input image $I_{LQ}\in\mathbb{R}^{H\times W\times C_{in}}$, we use one $3\times 3$ convolution layer $H_{SF}(\cdot)$ to extract the shallow feature $F_0\in\mathbb{R}^{H\times W\times C}$ as:
\begin{equation}
  F_0=H_{SF}(I_{LQ}),
\end{equation}
where $C_{in}$ and $C$ denote the channel number of the input and the intermediate feature, respectively.
The shallow feature extraction can simply map the input from low-dimensional space to high-dimensional space, while achieving the high-dimensional embedding for each pixel token. 
Moreover, the early convolution layer can help learn better visual representation~\cite{uniformer} and lead to stable optimization~\cite{vitc}. 
We then perform deep feature extraction $H_{DF}(\cdot)$ to further obtain the deep feature $F_{DF}\in\mathbb{R}^{H\times W\times C}$ as:
\begin{equation}
  F_{DF}=H_{DF}(F_0),
\end{equation}
where $H_{DF}(\cdot)$ consists of $N_1$ residual hybrid attention groups (RHAG) and one $3\times 3$ convolution layer $H_{Conv}(\cdot)$. 
These RHAGs progressively process the intermediate features as:
\begin{gather}
  F_{i}=H_{RHAG_i}(F_{i-1}), i=1,2,...,N, \notag \\ 
  F_{DF}=H_{Conv}(F_N), 
\end{gather}
where $H_{RHAG_i}(\cdot)$ represents the $i$-th RHAG. 
Following \cite{swinir}, we also introduce a convolution layer at the tail of this part to better aggregate information of deep features. 
After that, we add a global residual connection to fuse shallow features and deep features, and then reconstruct the high-quality (HQ) result via a reconstruction module as:
\begin{equation}
  I_{HQ}=H_{Rec}(F_0+F_{DF}),
\end{equation}
where $H_{Rec}(\cdot)$ denotes the reconstruction module. 
We adopt the pixel-shuffle method~\cite{pixelshuffle} to up-sample the fused feature for the SR task shown in Fig.~\ref{Network Structure}(a), and use two convolutions for the tasks where have the input and output have the same resolution shown in Fig.~\ref{Network Structure}(b). 
The key component RHAG consists of $N_2$ hybrid attention blocks (HAB), one overlapping cross-attention block (OCAB), one $3\times 3$ convolution layer, with a residual connection, as presented in Fig.~\ref{Network Structure}(c).

\subsection{Hybrid Attention Block} 
\label{HAB}
In this section, we detail our proposed Hybrid Attention Block (HAB), illustrated in Fig.~\ref{Network Structure}(d). HAB adopts a structure similar to the standard Swin Transformer block, preserving the window-based self-attention mechanism. However, we enhance the representative ability of self-attention and introduce a channel attention block to capture global information.
As discussed in Sec.\ref{Motivation}, we aim to activate more input pixels for Transformer to achieve stronger reconstruction capability. Unlike convolution that expands the receptive field by stacking layers, self-attention possesses a global receptive field within its scope. Therefore, a natural approach to expand the range of information utilized by window-based self-attention is to enlarge the window size. Previous work\cite{swinir} limits the window size used for self-attention calculations to a small range (i.e., 7 or 8), relying on a shifted window mechanism to gradually expand the receptive field. While this method reduces computational cost, it compromises the effectiveness of self-attention. As discussed in Sec.~\ref{cmp_win_size}, we find that window size is a crucial factor influencing the ability of window-based self-attention to exploit information. Appropriately increasing the window size can significantly improve the Transformer’s performance. Therefore, in HAB, we adopt a larger window size (i.e., 16).
In addition to window-based self-attention, global information can also be captured by incorporating channel attention. We think that global information may help for cases where many similar textures are present, as shown in Fig.\ref{cem}. Moreover, several studies\cite{restormer, nafnet} have demonstrated that channel-wise dynamic mapping is beneficial for low-level vision tasks. Therefore, we introduce the channel attention mechanism into our network. Given the evidence that convolution can improve the visual representation of Transformer models and facilitate easier optimization~\cite{cvt, vitc, ceit, uniformer, spach}, we incorporate a channel attention-based convolution block, referred to as the channel attention block (CAB), into the standard Transformer block to construct our HAB (see Fig.~\ref{Network Structure}(f)).
In addition, global information can be utilized when channel attention is adopted, as it is involved to calculate the channel attention weights.
We think that the global information may help for cases where many similar textures exist, such as the examples given in Fig.~\ref{lam}.
Besides, several works~\cite{restormer,nafnet} have demonstrated that the channel-wise dynamic mapping is beneficial for low-level vision tasks.
Therefore, we want to introduce the channel attention mechanism to our network.
Since many works have shown that convolution can help Transformer get better visual representation or achieve easier optimization~\cite{cvt,vitc,ceit,uniformer,spach}, we incorporate a channel attention-based convolution block, i.e., the channel attention block (CAB) in Fig.~\ref{Network Structure}(f), into the standard Transformer block to build our HAB.
To avoid the possible conflict of CAB and MSA on optimization and visual representation, we combine them in parallel and set a small constant $\alpha$ to control the weight of the CAB output. 
Overall, for a given input feature $X$, the whole process of HAB is computed as:
\begin{gather}
  X_N={\rm LN}(X), \notag \\
  X_M={\rm \text{(S)W-MSA}}(X_N)+\alpha {\rm CAB} (X_N)+X, \\
  Y={\rm MLP}({\rm LN}(X_M))+X_M, \notag 
\end{gather}
where $X_N$ and $X_M$ denote the intermediate features. $Y$ represents the output of HAB. 
LN represents the layer normalization operation and MLP denotes a multi-layer perceptron.
(S)W-MSA means the standard and shifted window multihead self-attention modules. 
Especially, we treat each pixel as a token for embedding (\textit{i.e.}, set patch size as 1 for patch embedding following~\cite{swinir}).  For calculation of the self-attention module, given an input feature of size $H\times W\times C$, it is first partitioned into $\frac{HW}{M^2}$ local windows of size $M\times M$, then self-attention is calculated inside each window. For a local window feature $X_W\in\mathbb{R}^{M^2\times C}$, the \textit{query}, \textit{key} and \textit{value} matrices are computed by linear mappings as $Q$, $K$ and $V$. Then the window-based self-attention is formulated as:
\begin{equation}
    {\rm Attention}(Q,K,V)={\rm SoftMax}(QK^T/\sqrt{d}+B)V, \label{attn}
\end{equation}
where $d$ represents the dimension of \textit{query}/\textit{key}. $B$ denotes the relative position encoding and is calculated as \cite{transformer}. Besides, to build the connections between neighboring non-overlapping windows, we also use the shifted window partitioning approach~\cite{swin_t}, with the shift size set to half of the window size. 

A CAB consists of two standard convolution layers with GELU activation~\cite{GELU} and a channel attention (CA) module, as shown in Fig.\ref{Network Structure}(f). Since Transformer-based structures often require a large number of channels for token embedding, directly using convolutions with constant width would result in high computational costs. To address this, we compress the number of channels in the two convolution layers by a constant factor $\beta$. For an input feature with $C$ channels, the number of channels is reduced to $\frac{C}{\beta}$ after the first convolution layer, and then expanded back to $C$ channels through the second layer. Finally, a standard CA module\cite{rcan} is employed to adaptively rescale channel-wise features.

\begin{figure}[!t]
\centering
\includegraphics[width=0.95\linewidth]{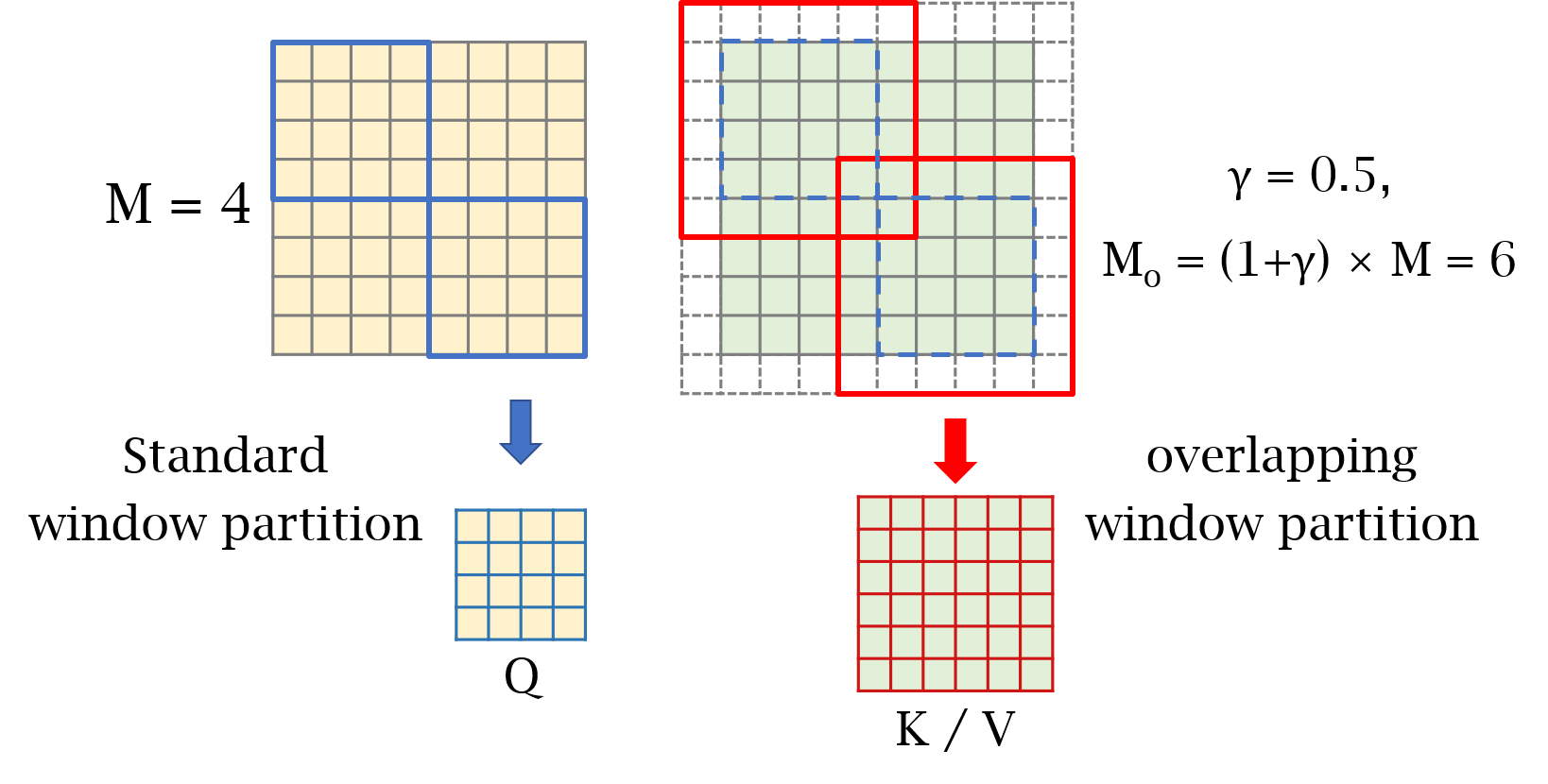}
\vspace{-3pt}
\caption{The overlapping window partition for OCA.}
\vspace{-9pt}
\label{OCA}
\end{figure}

\subsection{Overlapping Cross-Attention Block (OCAB)}
\label{ocab}
We introduce OCAB to directly establish cross-window connections and enhance the representative ability for the window self-attention. Our OCAB consists of an overlapping cross-attention (OCA) layer and an MLP layer similar to the standard Swin Transformer block~\cite{swin_t}.  But for OCA, as depicted in Fig.~\ref{OCA}, we use different window sizes to partition the projected features. Specifically, for the $X_Q,X_K,X_V\in\mathbb{R}^{H\times W\times C}$ of the input feature $X$, $X_Q$ is partitioned into $\frac{HW}{M^2}$ non-overlapping windows of size ${M}\times {M}$, while $X_K,X_V$ are unfolded to $\frac{HW}{M^2}$ overlapping windows of size ${M_o}\times {M_o}$. It is calculated as
\begin{equation}
    M_o=(1+\gamma)\times M,
\end{equation}
where $\gamma$ is a constant to control the overlapping size. To better understand this operation, the standard window partition can be considered as a sliding partition with the kernel size and the stride both equal to the window size $M$. In contrast, the overlapping window partition can be viewed as a sliding partition with the kernel size equal to $ M_o$, while the stride is equal to $M$. Zero-padding with size $\frac{\gamma M}{2}$ is used to ensure the size consistency of overlapping windows. The attention matrix is calculated as Eq.~\eqref{attn}, and the relative position bias $B\in\mathbb{R}^{M\times M_o}$ is also adopted. Unlike WSA whose \textit{query}, \textit{key} and \textit{value} are calculated from the same window feature, OCA computes \textit{key}/\textit{value} from a larger field where more useful information can be utilized for the \textit{query}. 
Note that although Multi-resolution Overlapped Attention (MOA) module in~\cite{moa} performs similar overlapping window partition, our OCA is fundamentally different from MOA. MOA calculates global attention using window features as tokens, while OCA computes cross-attention inside each window using pixel tokens. 

\subsection{The Same-task Pre-training}
Pre-training is proven effective on many high-level vision tasks~\cite{vit,beit,he2022masked}. Recent works~\cite{ipt,edt} also demonstrate that pre-training is beneficial to low-level vision tasks. IPT~\cite{ipt} emphasizes the use of various low-level tasks, such as denoising, deraining, SR and \textit{etc.}, while EDT~\cite{edt} utilizes different degradation levels of a specific task to do pre-training. These works focus on investigating the effect of multi-task pre-training for a target task. In contrast, we directly perform pre-training on a larger-scale dataset (\textit{i.e.}, ImageNet~\cite{imagenet}) based on the same task, showing that the effectiveness of pre-training depends more on the scale and diversity of data. For example, when we want to train a model for $\times 4$ SR, we first train a $\times 4$ SR model on ImageNet, then fine-tune it on the specific dataset, such as DF2K. The proposed strategy, namely \textit{same-task pre-training}, is simpler while bringing more performance improvements. It is worth mentioning that sufficient training iterations for pre-training and an appropriate small learning rate for fine-tuning are very important for the effectiveness of the pre-training strategy. We think that it is because Transformer requires more data and iterations to learn general knowledge for the task, but needs a small learning rate for fine-tuning to avoid overfitting to the specific dataset.

\subsection{Discussions}
In this part, we analyze the distinctions of our HAT and 
several relevant works, including SwinIR~\cite{swinir}, 
EDT~\cite{edt}, SCUNet~\cite{scunet} and HaloNet~\cite{halonet}.

\textbf{Difference to SwinIR.} SwinIR~\cite{swinir} is the first work to successfully use Swin Transformer~\cite{swin_t} for low-level vision tasks. It builds an image restoration network by using the original Swin Transformer block. Our HAT is inspired by SwinIR and retains the core design of window-based self-attention. However, we address the problem of limited range of utilized information in SwinIR by enlarging the window size and introducing channel attention. At the same time, we introduce a newly designed OCA to further enhance the ability to implement cross-window interaction. This work aims to design a more powerful backbone for image restoration tasks.


\textbf{Difference to EDT.} EDT~\cite{edt} builds an image restoration Transformer based on the shifted crossed local attention, which also calculates self-attention in the windows of fixed size. For a given feature map, it splits the feature into two parts and performs self-attention in an either horizontal or vertical rectangle window. In contrast, our HAT adopts the vanilla window-based self-attention and shifted window mechanism similar to Swin Transformer~\cite{swin_t}. EDT also studies the pre-training strategy and emphasizes the advantages of multi-related-task pre-training (i.e., performing pre-training on a specific task with multiple degradation levels). However, HAT shows that training on the same task but using a large-scale dataset is the key factor in the effectiveness of pre-training. 

\textbf{Difference to SCUNet.} SCUNet~\cite{scunet} is also an image restoration network that integrates the strengths of Transformers and CNN. It utilizes the Swin Transformer block alongside a classic convolution block within its U-Net architecture, forming a Swin-Conv Block, and achieves excellent performance for denoising. Unlike our approach that originates from the SR task, SCUNet is primarily designed for denoising, with a focus on capturing multi-scale information. In contrast, our method emphasizes the benefits of window self-attention for local information fitting and addresses its limitations in cross-window interaction and global information acquisition. Therefore, the two methods differ significantly in motivation, overall architecture and the design details of key modules.

\textbf{Difference to HaloNet.} HaloNet~\cite{halonet} incorporates a similar window partition mechanism to our OCA, enabling the calculation of self-attention within overlapping window features. HaloNet employs this overlapping self-attention as the fundamental module to build the network, inevitably leading to a large computational cost. This design could impose a substantial computational burden and is not friendly to image restoration tasks. On the contrary, our HAT leverages only a limited number of OCA modules to augment the interaction between adjacent windows. This approach can effectively enhance the image restoration Transformer without incurring excessive computational costs.

\subsection{Implementation Details}
For the structure of HAT, both the RHAG number and HAB number are set to 6. The channel number of the whole network is set to 180. The attention head number and window size are set to 6 and 16 for both (S)W-MSA and OCA. For the specific hyper-parameters of the proposed modules, we set the weighting factor of CAB output ($\alpha$), the squeeze factor between two convolution layers in CAB ($\beta$), and the overlapping ratio of OCA ($\gamma$) as 0.01, 3 and 0.5, respectively. For the large variant HAT-L, we double the depth of HAT by increasing the RHAG number from 6 to 12. We also provide a small version HAT-S with fewer parameters and similar computation to SwinIR. For HAT-S, we set the channel number to 144 and set $\beta$ to 24 in CAB. When implementing the pre-training strategy, we adopt ImageNet~\cite{imagenet} as the pre-training dataset following \cite{ipt,edt}. We conduct the main experiments and ablation study on image SR. Therefore, we use the DF2K dataset (DIV2K~\cite{div2k}+Flicker2K~\cite{flicker2k}) as the training dataset, following~\cite{swinir,rcanit}. PSNR/SSIM calculated on the Y channel is reported for the quantitative metrics.

\begin{table}[!t]
\center
\begin{center}
\caption{Performance and complexity for different window sizes.}
\vspace{-0.1cm}
\label{window_analysis}
\setlength{\tabcolsep}{1.6mm}{
\begin{tabular}{c|cc|ccc} 
\hline 
Window Size & Params. & Multi-Adds. & Set5 & Set14 & Urban100 \\
\hline 
(8,8) & 11.9M & 53.6G & 32.88dB & 29.09dB & 27.45dB \\
(16,16) & 12.1M & 63.8G & 32.97dB & 29.12dB & 27.81dB \\
\hline 
\end{tabular}
}
\end{center}
\vspace{-0.2cm}
\end{table}

\begin{figure}[!t]
\centering
\includegraphics[width=0.95\linewidth]{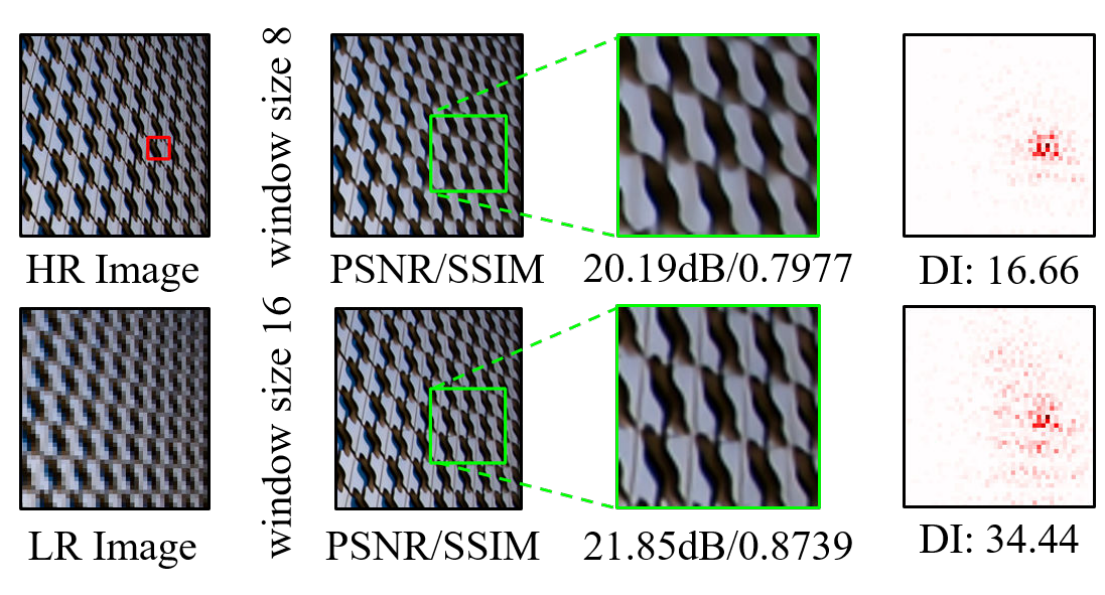}
\vspace{-0.2cm}
\caption{Qualitative comparison of different window sizes.}
\vspace{-0.4cm}
\label{win_size_cmp_fig}
\end{figure}

\section{Network Investigation}
\subsection{Effects of Different Window Sizes}
\label{cmp_win_size}
As discussed in Sec. \ref{Motivation}, activating more input pixels for the restoration tends to achieve better performance. Enlarging window size for the window-based self-attention is an intuitive way to realize the goal. To examine how the window size in self-attention affects both performance and computational efficiency of Transformer models for image SR, we conduct a series of experiments directly on the preliminary version of SwinIR, without involving newly-introduced blocks (e.g., OCAB and CAB).
As shown in Tab.~\ref{window_analysis}, enlarging the window size from $8 \times 8$ to $16 \times 16$ consistently improves performance across all benchmark datasets. Especially on Urban100, we observe a significant gain of 0.36dB (from 27.45dB to 27.81dB). We also provide the qualitative comparison in Fig.~\ref{win_size_cmp_fig}. The result produced by the model with a larger window size has much clearer textures. For LAM results, the model with window size of 16 utilizes much more input pixels than the model with window size of 8.
Meanwhile, the increase in computational cost is moderate. The model size grows slightly (from 11.9M to 12.1M parameters), and Multi-Adds (counted at the input size of 64 $\times$ 64) increase by $\sim$$\%19$ (from 53.6G to 63.8G). This suggests that enlarging window size is a highly cost-effective design for improving Transformer-based SR.

\subsection{Ablation Study}
\begin{figure}[!t]
\centering
\includegraphics[width=0.97\linewidth]{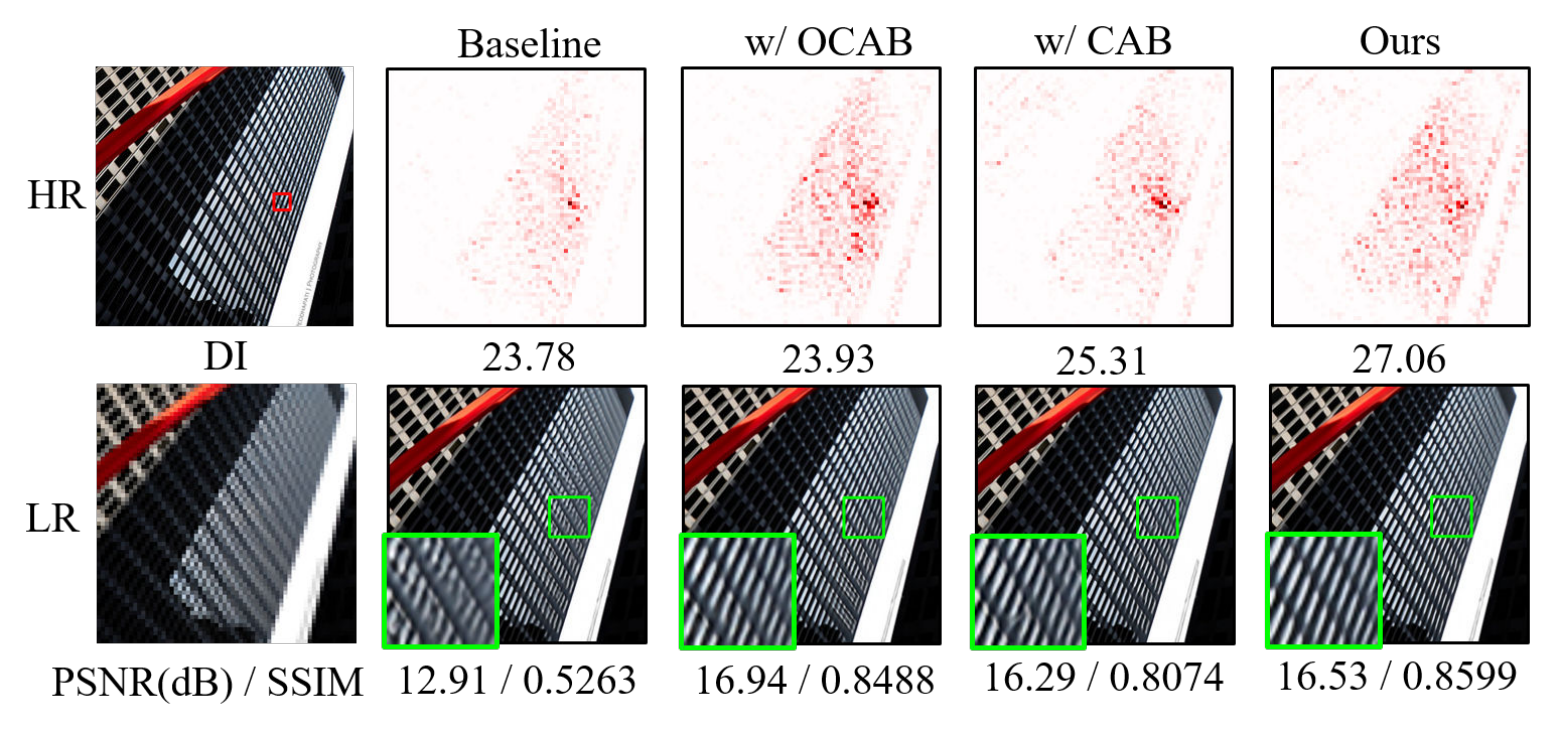}
\vspace{-0.2cm}
\caption{Ablation study on the proposed OCAB and CAB. 
}
\vspace{-0.4cm}
\label{ablation_study_fig}
\end{figure}

\textbf{Effectiveness of OCAB and CAB.}
To verify the effectiveness of the proposed CAB and OCAB. we conduct ablation study and complexity analysis based on $\times4$ SR and show the results in Tab.~\ref{ablation_study_tab}.
On Urban100, compared with the baseline results, both OCAB and CAB bring the performance gain of 0.1dB. Benefiting from the two modules, the model obtains a further performance gain of 0.16dB. On Set5 and Set14, the proposed OCAB and CAB can also bring considerable performance improvement. 
Besides, we investigate the computational complexity of OCAB and CAB. While OCAB introduces a modest increase in parameters and Multi-Adds, CAB is more computationally expensive. However, both modules bring stable and significant improvements.
We think that the performance improvement comes from two aspects. On the one hand, improving the window interaction by OCAB and utilizing the global statistics by CAB both help the model better deal with the long-term patters (e.g., self-similarity of repeated textures). On the other hand, the two modules enrich and enhance the model ability by introducing cross-attention and convolution blocks.
We also provide qualitative comparison to further illustrate the influence of OCAB and CAB, as presented in Fig.~\ref{ablation_study_fig}. We can observe that the model with OCAB has a larger scope of the utilized pixels and generate better-reconstructed results. 
When CAB is adopted, the used pixels even expand to almost the full image. Moreover, the result of our method with OCAB and CAB obtains the highest DI\cite{lam}, which means our method utilizes the most input pixels. 

\begin{table}[!t]
\center
\begin{center}
\caption{Ablation study on the proposed OCAB and CAB.}
\vspace{-0.1cm}
\label{ablation_study_tab}
\setlength{\tabcolsep}{3.5mm}{
\begin{tabular}{c|cccc} 
\hline 
~ & \multicolumn{4}{c}{Baseline} \\
\hline 
OCAB & \text{\sffamily X} & \checkmark & \text{\sffamily X} & \checkmark\\
CAB & \text{\sffamily X} & \text{\sffamily X} & \checkmark & \checkmark\\
\hline 
Params. & 12.1M & 13.7M & 19.2M & 20.8M\\
Multi-Adds. & 63.8G & 74.7G & 92.8G & 103.7G\\
\hline 
Set5 & 32.97dB & 33.02dB & 33.00dB & 33.04dB\\
Set14 & 29.12dB & 29.19dB & 29.16dB & 29.23dB\\
Urban100 & 27.81dB & 27.91dB & 27.91dB & 27.97dB\\
\hline 
\end{tabular}
}
\end{center}
\vspace{-0.2cm}
\end{table}

\noindent
\textbf{Effects of different designs of CAB.} 
We conduct experiments to explore the effects of different designs of CAB. The results are reported on the Urban100 dataset. First, we investigate the influence of channel attention. As shown in Tab. \ref{CAB_structure}, the model using CA achieves a performance gain of 0.05dB compared to the model without CA, demonstrating the effectiveness of the CA in our network. We also conduct experiments to explore the effects of the weighting factor $\alpha$ of CAB. As presented in Sec. \ref{HAB}, $\alpha$ is used to control the weight of CAB features for feature fusion. A larger $\alpha$ means a larger weight of features extracted by CAB and $\alpha=0$ represents CAB is not used. As shown in Tab.~\ref{weighting_factor}, the model with $\alpha=0.01$ obtains the best performance. It indicates that CAB and self-attention may have potential optimization conflict, while a small weighting factor for our CAB can suppress this issue for better combination.

\noindent
\textbf{Effects of the overlapping ratio.}
In OCAB, we set a constant $\gamma$ to control the overlapping size for the overlapping cross-attention, as illustrated in Sec \ref{ocab}. To explore the effects of different overlapping ratios, we set a group of $\gamma$ from 0 to 0.75 to examine the performance change, as shown in Tab. \ref{overlapping size}. Note that $\gamma=0$ means a standard Transformer block. It can be found that the model with $\gamma=0.5$ performs best. In contrast, when $\gamma$ is set to 0.25 or 0.75, the model has no obvious performance gain or even has a performance drop. It illustrates that inappropriate overlapping size cannot benefit the interaction of neighboring windows. 

\subsection{Analysis of Model Complexity}
\label{complexity}
We analyze the computational complexity of our method from two perspectives: (1) the impact of different CAB sizes, and (2) a comparison between our method and SwinIR under similar computational budgets. All experiments are conducted on Urban100 for $\times$4 SR, with the number of Multiply-Add operations counted at an input size of $64 \times 64$. Pre-training is not used in any model, and all results are obtained under identical training settings.
Since CAB seems to be computationally expensive, we first investigate the influence of the squeeze factor $\beta$ in CAB (mentioned in Sec \ref{HAB}), which controls the channel reduction ratio. As shown in Tab.~\ref{cab_cmp}, adding a small CAB whose $\beta$ equals 6 can bring considerable performance improvement. When we continuously reduce $\beta$, the performance increases but with larger model sizes. To balance the performance and computations, we set $\beta$ to 3 as the default setting. 

\begin{table}[!t]
\center
\begin{center}
\caption{Effects of the channel attention (CA) module in CAB.}
\vspace{-0.1cm}
\label{CAB_structure}
\setlength{\tabcolsep}{2.5mm}{
\begin{tabular}{c|cc} 
\hline 
Structure & w/o CA & w/ CA\\
\hline 
PSNR / SSIM & 27.92dB / 0.8362 & 27.97dB / 0.8367\\
\hline 
\end{tabular}
}
\end{center}
\end{table}

\begin{table}[!t]
\center
\begin{center}
\caption{Effects of the weighting factor $\alpha$ in CAB.}
\vspace{-0.1cm}
\label{weighting_factor}
\setlength{\tabcolsep}{2.5mm}{
\begin{tabular}{c|cccc} 
\hline 
$\alpha$ & 0 & 1 & 0.1 & 0.01\\
\hline
PSNR & 27.81dB & 27.86dB & 27.90dB & 27.97dB\\
\hline 
\end{tabular}
}
\end{center}
\end{table}

\begin{table}[!t]
\center
\begin{center}
\caption{Ablation study on the overlapping ratio of OCAB.}
\vspace{-0.1cm}
\label{overlapping size}
\setlength{\tabcolsep}{2.5mm}{
\begin{tabular}{c|cccc} 
\hline 
$\gamma$ & 0 & 0.25 & 0.5 & 0.75\\
\hline 
PSNR & 27.85dB & 27.81dB & 27.91dB & 27.86dB\\
\hline 
\end{tabular}
}
\end{center}
\end{table}

\begin{table}[!t]
\center
\begin{center}
\caption{Model complexity for different CAB sizes.}
\vspace{-0.1cm}
\label{cab_cmp}
\setlength{\tabcolsep}{2.5mm}{
\begin{tabular}{c|ccc} 
\hline 
$\beta$ in CAB & Params. & Multi-Adds. & PSNR\\
\hline 
1 & 33.2M & 150.1G & 27.97dB \\
\hline 
2 & 22.7M & 107.1G & 27.92dB \\
\hline 
3 (default) & 19.2M & 92.8G & 27.91dB \\
\hline 
6 & 15.7M & 78.5G & 27.88dB \\
\hline 
w/o CAB & 12.1M & 63.8G & 27.81dB \\
\hline 
\end{tabular}
}
\end{center}
\vspace{-0.1cm}
\end{table}

\begin{table}[!t]
\center
\begin{center}
\caption{Model complexity comparison of SwinIR and HAT.}
\vspace{-0.1cm}
\label{method_cmp}
\setlength{\tabcolsep}{2.5mm}{
\begin{tabular}{c|ccc} 
\hline 
Method & Params. & Multi-Adds. & PSNR\\
\hline 
SwinIR & 11.9M & 53.6G & 27.45dB \\
\hline 
HAT-S (ours) & 9.6M & 54.9G & 27.80dB \\
\hline 
\hline 
SwinIR-L1 & 24.0M & 104.4G & 27.53dB \\
\hline 
SwinIR-L2 & 23.1M & 102.4G & 27.58dB \\
\hline 
HAT (ours) & 20.8M & 103.7G & 27.97dB \\
\hline 
\end{tabular}
}
\end{center}
\vspace{-0.1cm}
\end{table}

To further validate the efficiency of our method, we compare HAT and SwinIR with the similar numbers of parameters and Multi-Adds under two settings, as presented in Tab.~\ref{method_cmp}. First, we compare HAT-S with the original version of SwinIR. With less parameters and comparable computations, HAT-S significantly outperforms SwinIR. Second, we enlarge SwinIR by increasing the width and depth to achieve similar computations to HAT, denoted as SwinIR-L1 and SwinIR-L2.
HAT achieves the best performance at the lowest computational cost. This demonstrates that HAT outperforms SwinIR in performance and computational efficiency.

Overall, Our experiments show that properly tuning the CAB size (via $\beta$) enables flexible trade-offs between performance and complexity. Compared with SwinIR, our HAT achieves higher performance under similar or lower computational costs, in both compact and large settings. These results validate the efficiency and scalability of our proposed method.

\begin{table}[!t]
\center
\begin{center}
\caption{Quantitative results of HAT using two kinds of pre-training strategies on $\times$4 SR under the same training setting. 
}
\label{pretrain1}
\setlength{\tabcolsep}{1.8mm}{
\begin{tabular}{c|c|ccc} 
\hline 
Strategy & Stage & Set5 & Set14 & Urban100 \\
\hline 
Multi-related-task & pre-training & 32.94dB & 29.17dB & 28.05dB\\
pre-training & fine-tuning & 33.06dB & 29.33dB & 28.21dB\\
\hline 
Same-task & pre-training & 33.02dB & 29.20dB & 28.11dB\\
pre-training(ours) & fine-tuning & 33.07dB & 29.34dB & 28.28dB\\
\hline 
\end{tabular}
}
\end{center}
\vspace{-0.3cm}
\end{table}

\begin{figure}[!t]
	\begin{center}
    \includegraphics[width=1\linewidth]{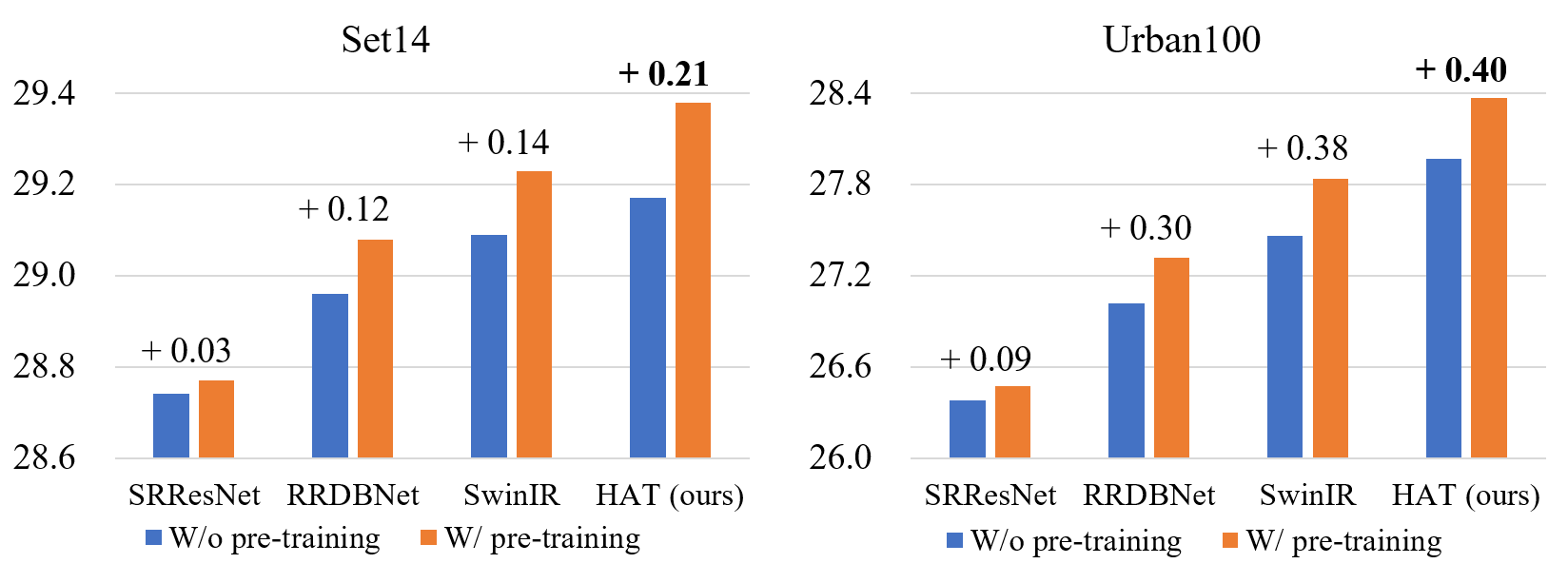}
	\end{center}
    \vspace{-0.2cm}
	\caption{Quantitative comparison on PSNR(dB) of four different networks without and with the same-task pre-training on $\times$4 SR.}
    \label{pretrain2}
    \vspace{-0.3cm}
\end{figure}

\subsection{Study on the Pre-training Strategy}
\label{cmp_pretrain}
From Tab.~\ref{quantitative results}, we can see that HAT can benefit greatly from the pre-training strategy, by comparing the performance of HAT and {HAT}$^\dagger$. To show the superiority of the proposed same-task pre-training, we also apply the multi-related-task pre-training~\cite{edt} to HAT for comparison using full ImageNet, under the same training settings as ~\cite{edt}. As depicted as Tab.~\ref{pretrain1}, the same-task pre-training performs better, not only in the pre-training stage but also in the fine-tuning process. From this perspective, multi-task pre-training probably impairs the restoration performance of the network on a specific degradation, while the same-task pre-training can maximize the performance gain brought by large-scale data. To further investigate the influences of our pre-training strategy for different networks, we apply our pre-training to four networks: SRResNet (1.5M), RRDBNet (16.7M), SwinIR (11.9M) and HAT (20.8M), as shown in Fig.~\ref{pretrain2}. First, we can see that all four networks can benefit from pre-training, showing the effectiveness of the proposed same-task pre-training strategy. Second, for the same type of network (\textit{i.e.}, CNN or Transformer), the larger the network capacity, the more performance gain from pre-training. Third, although with less parameters, SwinIR obtains greater performance 
improvement from the pre-training compared to RRDBNet. It suggests that Transformer needs more data to exploit the potential of the model. HAT obtains the largest gain from pre-training, indicating the necessity of the pre-training strategy for such large models. Equipped with big models and large-scale data, we show that the performance upper bound of this task can be significantly extended.

\begin{table*}[!ht]
\centering
\caption{Quantitative comparison with state-of-the-art methods for \textbf{\underline{classic image super-resolution}} on benchmark datasets. The best and second best results are marked in \textbf{bold} and \underline{underline}. ``$\dagger$'' indicates that methods adopt pre-training strategy on ImageNet.}
\vspace{-0.3cm}
\label{quantitative results}

\renewcommand{\arraystretch}{1.1}
\resizebox{\textwidth}{88mm}{
\begin{tabular}{|l|c|c|c|c|c|c|c|c|c|c|c|c|}
\hline
\multirow{2}{*}{Method} & \multirow{2}{*}{Scale} & \multirow{2}{*}{\makecell{Training\\Dataset}} &
\multicolumn{2}{c|}{Set5~\cite{set5}} &  \multicolumn{2}{c|}{Set14~\cite{set14}} &  \multicolumn{2}{c|}{BSD100~\cite{bsd100}} &  \multicolumn{2}{c|}{Urban100~\cite{urban100}} &  \multicolumn{2}{c|}{Manga109~\cite{manga109}}  
\\ 
\cline{4-13}
&  &  & PSNR & SSIM & PSNR & SSIM & PSNR & SSIM & PSNR & SSIM & PSNR & SSIM 
\\ 
\hline
\hline
EDSR~\cite{edsr} & $\times$2 & DIV2K %
& 38.11
& 0.9602
& 33.92
& 0.9195
& 32.32
& 0.9013
& 32.93
& 0.9351
& 39.10
& 0.9773
\\
RCAN~\cite{rcan} & $\times$2 & DIV2K %
& 38.27
& 0.9614
& 34.12
& 0.9216
& 32.41
& 0.9027
& 33.34
& 0.9384
& 39.44
& 0.9786
\\  
SAN~\cite{san} & $\times$2 & DIV2K %
& {38.31}
& {0.9620}
& {34.07}
& {0.9213}
& {32.42}
& {0.9028}
& {33.10}
& {0.9370}
& {39.32}
& {0.9792}\\
IGNN~\cite{ignn} & $\times$2 & DIV2K %
& {38.24}
& {0.9613}
& {34.07}
& {0.9217}
& {32.41}
& {0.9025}
& {33.23}
& {0.9383}
& {39.35}
& {0.9786}
\\
HAN~\cite{han} & $\times$2 & DIV2K %
& {38.27}
& {0.9614}
& {34.16}
& {0.9217}
& {32.41}
& {0.9027}
& {33.35}
& {0.9385}
& {39.46}
& {0.9785}              
\\
RCAN-it~\cite{rcanit} & $\times$2 & DF2K %
& 38.37
& 0.9620
& 34.49
& 0.9250
& 32.48
& 0.9034
& 33.62
& 0.9410
& 39.88
& 0.9799
\\
SwinIR~\cite{swinir} & $\times$2 & DF2K %
& 38.42
& 0.9623
& 34.46
& 0.9250
& 32.53
& 0.9041
& 33.81
& 0.9427
& 39.92
& 0.9797
\\
EDT~\cite{edt} & $\times$2 & DF2K %
& 38.45
& 0.9624
& 34.57
& 0.9258
& 32.52
& 0.9041
& 33.80
& 0.9425
& 39.93
& 0.9800
\\
\textbf{HAT-S} (ours) & $\times$2 & DF2K %
& {38.58}
& {0.9628}
& {34.70}
& {0.9261}
& {32.59}
& {0.9050}
& {34.31}
& {0.9459}
& {40.14}
& {0.9805}
\\
\textbf{HAT} (ours) & $\times$2 & DF2K %
& {38.63}
& {0.9630}
& {34.86}
& {0.9274}
& {32.62}
& {0.9053}
& {34.45}
& {0.9466}
& {40.26}
& {0.9809}
\\
\hdashline
IPT$^\dagger$~\cite{ipt} & $\times$2 & ImageNet %
& {38.37}
& {-}
& {34.43}
& {-}
& {32.48}
& {-}
& {33.76}
& {-}
& {-}
& {-}
\\
EDT$^\dagger$~\cite{edt} & $\times$2 & DF2K %
& {38.63}
& {0.9632}
& 34.80
& 0.9273
& {32.62}
& {0.9052}
& 34.27
& 0.9456
& {40.37}
& {0.9811}
\\
GRL$^\dagger$~\cite{grl} & $\times$2 & Unknown %
& 38.67
& \underline{0.9640}
& 35.08
& \underline{0.9288}
& 32.67
& \underline{0.9065}
& \underline{35.06}
& \textbf{0.9505}
& 40.67
& \underline{0.9824}
\\
\textbf{HAT}$^\dagger$ (ours) & $\times$2 & DF2K %
& \underline{38.73}
& 0.9637
& \underline{35.13}
& 0.9282
& \underline{32.69}
& 0.9060
& 34.81
& \underline{0.9489}
& \underline{40.71}
& 0.9819
\\
\textbf{HAT-L}$^\dagger$ (ours) & $\times$2 & DF2K %
& \textbf{38.91}
& \textbf{0.9646}
& \textbf{35.29}
& \textbf{0.9293}
& \textbf{32.74}
& \textbf{0.9066}
& \textbf{35.09}
& \textbf{0.9505}
& \textbf{41.01}
& \textbf{0.9831}
\\
\hline
\hline
EDSR~\cite{edsr} & $\times$3 & DIV2K %
& 34.65
& 0.9280
& 30.52
& 0.8462
& 29.25
& 0.8093
& 28.80
& 0.8653
& 34.17
& 0.9476
\\
RCAN~\cite{rcan} & $\times$3 & DIV2K %
& 34.74
& 0.9299
& 30.65
& 0.8482
& 29.32
& 0.8111
& 29.09
& 0.8702
& 34.44
& 0.9499
\\
SAN~\cite{san} & $\times$3 & DIV2K %
& {34.75}
& {0.9300}
& {30.59}
& {0.8476}
& {29.33}
& {0.8112}
& {28.93}
& {0.8671}
& {34.30}
& {0.9494}
\\
IGNN~\cite{ignn} & $\times$3 & DIV2K %
& {34.72}
& {0.9298}
& {30.66}
& {0.8484}
& {29.31}
& {0.8105}
& {29.03}
& {0.8696}
& {34.39}
& {0.9496}
\\
HAN~\cite{han}  & $\times$3 & DIV2K %
& {34.75}
& {0.9299}
& {30.67}
& {0.8483}
& {29.32}
& {0.8110}
& {29.10}
& {0.8705}
& {34.48}
& {0.9500}
\\
RCAN-it~\cite{rcanit} & $\times$3 & DF2K %
& {34.86}
& {0.9308}
& {30.76}
& {0.8505}
& {29.39}
& {0.8125}
& {29.38}
& {0.8755}
& {34.92}
& {0.9520}
\\
SwinIR~\cite{swinir} & $\times$3 & DF2K %
& 34.97
& 0.9318
& 30.93
& 0.8534
& 29.46
& 0.8145
& 29.75
& 0.8826
& 35.12
& 0.9537
\\
EDT~\cite{edt} & $\times$3 & DF2K %
& 34.97
& 0.9316
& 30.89
& 0.8527
& 29.44
& 0.8142
& 29.72
& 0.8814
& 35.13
& 0.9534
\\
\textbf{HAT-S} (ours) & $\times$3 & DF2K %
& {35.01}
& {0.9325}
& {31.05}
& {0.8550}
& {29.50}
& {0.8158}
& {30.15}
& {0.8879}
& {35.40}
& {0.9547}
\\
\textbf{HAT} (ours) & $\times$3 & DF2K %
& {35.07}
& {0.9329}
& {31.08}
& {0.8555}
& {29.54}
& {0.8167}
& {30.23}
& {0.8896}
& {35.53}
& {0.9552}
\\
\hdashline
IPT$^\dagger$~\cite{ipt} & $\times$3 & ImageNet %
& {34.81}
& {-}
& {30.85}
& {-}
& {29.38}
& {-}
& {29.49}
& {-}
& {-}
& {-}
\\
EDT$^\dagger$~\cite{edt} & $\times$3 & DF2K %
& {35.13}
& 0.9328
& {31.09}
& 0.8553
& {29.53}
& {0.8165}
& 30.07
& 0.8863
& 35.47
& 0.9550
\\
\textbf{HAT}$^\dagger$ (ours) & $\times$3 & DF2K %
& \underline{35.16}
& \underline{0.9335}
& \underline{31.33}
& \underline{0.8576}
& \underline{29.59}
& \underline{0.8177}
& \underline{30.70}
& \underline{0.8949}
& \underline{35.84}
& \underline{0.9567}
\\
\textbf{HAT-L}$^\dagger$ (ours) & $\times$3 & DF2K %
& \textbf{35.28}
& \textbf{0.9345}
& \textbf{31.47}
& \textbf{0.8584}
& \textbf{29.63}
& \textbf{0.8191}
& \textbf{30.92}
& \textbf{0.8981}
& \textbf{36.02}
& \textbf{0.9576}
\\
\hline
\hline
EDSR~\cite{edsr} & $\times$4 & DIV2K %
& 32.46
& 0.8968
& 28.80
& 0.7876
& 27.71
& 0.7420
& 26.64
& 0.8033
& 31.02
& 0.9148
\\
RCAN~\cite{rcan} & $\times$4 & DIV2K %
& 32.63
& 0.9002
& 28.87
& 0.7889
& 27.77
& 0.7436
& 26.82
& 0.8087
& 31.22
& 0.9173
\\
SAN~\cite{san} & $\times$4 & DIV2K %
& {32.64}
& {0.9003}
& {28.92}
& {0.7888}
& {27.78}
& {0.7436}
& {26.79}
& {0.8068}
& {31.18}
& {0.9169}
\\
IGNN~\cite{ignn} & $\times$4 & DIV2K %
& {32.57}
& {0.8998}
& {28.85}
& {0.7891}
& {27.77}
& {0.7434}
& {26.84}
& {0.8090}
& {31.28}
& {0.9182}
\\
HAN~\cite{han} & $\times$4 & DIV2K %
& {32.64}
& {0.9002}
& {28.90}
& {0.7890}
& {27.80}
& {0.7442}
& {26.85}
& {0.8094}
& {31.42}
& {0.9177}
\\
RRDB~\cite{esrgan} & $\times$4 & DF2K %
& {32.73}
& {0.9011 }
& {28.99}
& {0.7917}
& {27.85}
& {0.7455}
& {27.03}
& {0.8153}
& {31.66}
& {0.9196}
\\
RCAN-it~\cite{rcanit} & $\times$4 & DF2K %
& 32.69
& 0.9007
& 28.99
& 0.7922
& 27.87
& 0.7459
& 27.16
& 0.8168
& 31.78
& 0.9217
\\
SwinIR~\cite{swinir} & $\times$4 & DF2K %
& 32.92
& 0.9044
& 29.09
& 0.7950
& 27.92
& 0.7489
& 27.45
& 0.8254
& 32.03
& 0.9260
\\
EDT~\cite{edt} & $\times$4 & DF2K %
& 32.82
& 0.9031
& 29.09
& 0.7939
& 27.91
& 0.7483
& 27.46
& 0.8246
& 32.05
& 0.9254
\\
\textbf{HAT-S} (ours) & $\times$4 & DF2K %
& {32.92}
& {0.9047}
& {29.15}
& {0.7958}
& {27.97}
& {0.7505}
& {27.87}
& {0.8346}
& {32.35}
& {0.9283}
\\
\textbf{HAT} (ours) & $\times$4 & DF2K %
& {33.04}
& {0.9056}
& {29.23}
& {0.7973}
& {28.00}
& {0.7517}
& {27.97}
& {0.8368}
& {32.48}
& {0.9292}
\\
\hdashline
IPT$^\dagger$~\cite{ipt} & $\times$4 & ImageNet %
& {32.64}
& {-}
& {29.01}
& {-}
& {27.82}
& {-}
& {27.26}
& {-}
& {-}
& {-}
\\
EDT$^\dagger$~\cite{edt} & $\times$4 & DF2K %
& {33.06}
& {0.9055}
& {29.23}
& {0.7971}
& {27.99}
& 0.7510
& 27.75
& 0.8317
& 32.39
& 0.9283
\\
GRL$^\dagger$~\cite{grl} & $\times$4 & Unknown %
& 33.10
& 0.9070
& 29.37
& \underline{0.8001}
& 28.01
& \underline{0.7538}
& \underline{28.53}
& \underline{0.8492}
& 32.77
& \underline{0.9325}
\\
\textbf{HAT}$^\dagger$ (ours) & $\times$4 & DF2K %
& \underline{33.18}
& \underline{0.9073}
& \underline{29.38}
& \underline{0.8001}
& \underline{28.05}
& 0.7534
& 28.37
& 0.8447
& \underline{32.87}
& 0.9319
\\
\textbf{HAT-L}$^\dagger$ (ours) & $\times$4 & DF2K %
& \textbf{33.30}
& \textbf{0.9083}
& \textbf{29.47}
& \textbf{0.8015}
& \textbf{28.09}
& \textbf{0.7551}
& \textbf{28.60}
& \textbf{0.8498}
& \textbf{33.09}
& \textbf{0.9335}
\\
\hline
\end{tabular}
}

\begin{minipage}{\textwidth}
\footnotesize
\vspace{0.1cm}
\emph{Note: For fair comparison, GRL's SSIM values are recalculated using BasicSR based on its official results, as the original reported values differ significantly.} 
\end{minipage}

\vspace{-0.3cm}
\end{table*}

\section{Experimental Results}
\subsection{Training Settings}
For classic image super-resolution, we use DF2K (DIV2K + Flicker2K) with 3450 images as the training dataset when training from scratch. The low-resolution images are generated from the ground truth images by the ``bicubic'' down-sampling in MATLAB. We set the input patch size to $64\times 64$ and use random rotation and horizontally flipping for data augmentation. The mini-batch size is set to 32 and total training iterations are set to 500K. The learning rate is initialized as 2e-4 and reduced by half at [250K,400K,450K,475K]. For $\times$4 SR, we initialize the model with pre-trained $\times$2 SR weights and halve the iterations for each learning rate decay as well as total iterations. We adopt Adam optimizer with $\beta_1=0.9$ and $\beta_2=0.99$ to train the model. When using the same-task pre-training, we exploit the full ImageNet dataset with 1.28 million images to pre-train the model for 800K iterations. The initial learning rate is also set to 2e-4 but reduced by half at [300K,500K,650K,700K,750k]. Then, we adopt DF2K dataset to fine-tune the pre-trained model. For fine-tuning, we set the initial learning rate to 1e-5 and halve it at 
[125K,200K,230K,240K] for total 250K iterations. 

For real-world image super-resolution, we train HAT models based on two simulated real-world degradation models, i.e., BSRGAN~\cite{bsrgan} and Real-ESRGAN~\cite{realesrgan}. 
The total batch size is set to 32 and the input patch size is set to $64\times 64$. The network structure is the same as the basic version of HAT for classic image super-resolution. Follwing Real-ESRGAN~\cite{realesrgan}, we first train the MSE-based model and introduce the generative adversarial training to fine-tune the GAN-based model. More training and degradation details can refer to \cite{bsrgan} and \cite{realesrgan}.

\begin{figure*}[!t]
	\centering
    \includegraphics[width=0.99\textwidth]{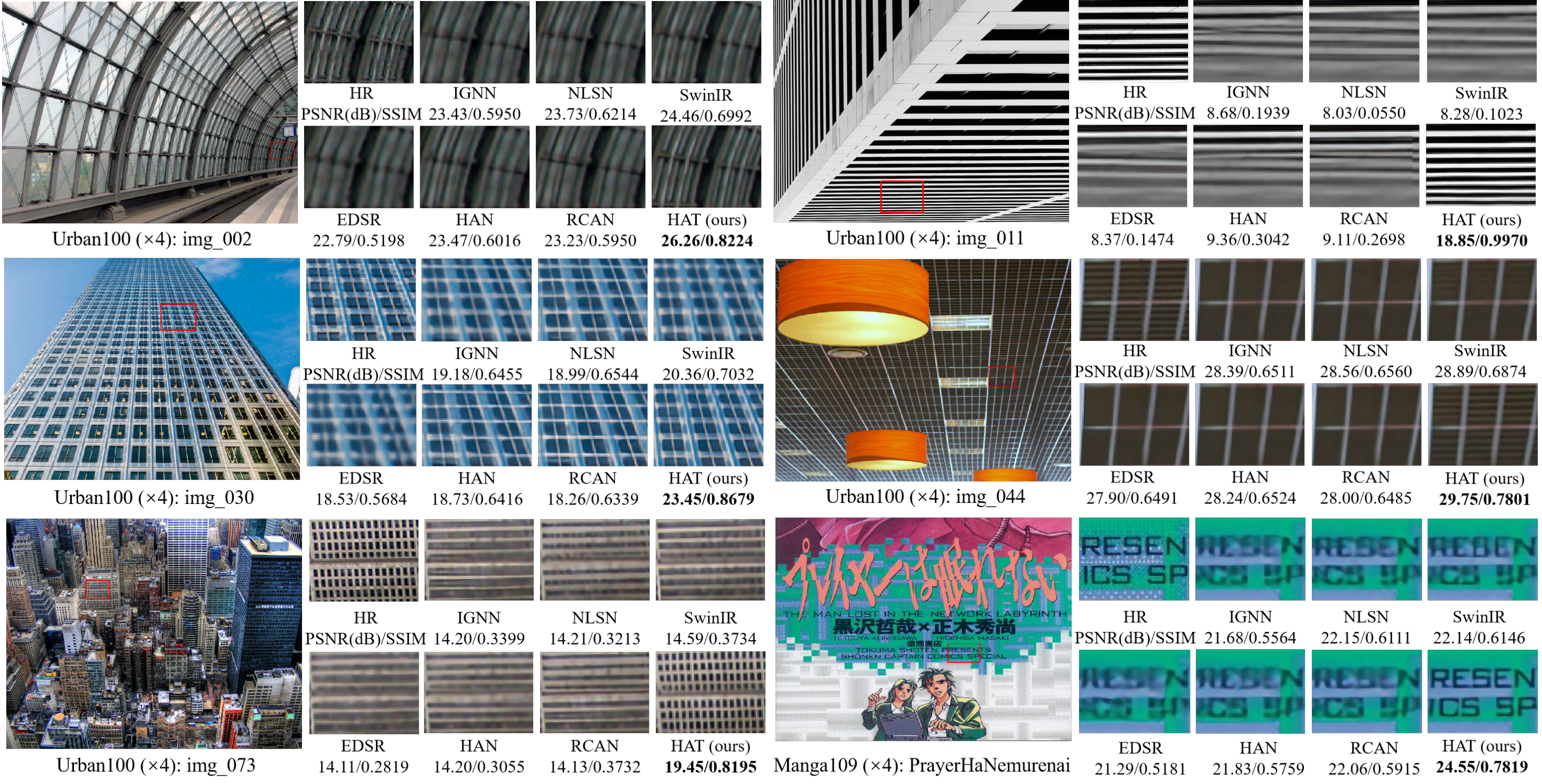}
    \vspace{-0.2cm}
	\caption{Visual comparison for $\times$4 SR. The patches for comparison are marked with red boxes in the original images. PSNR/SSIM is calculated based on the patches to better reflect the performance difference.}
    \vspace{-0.3cm}
    \label{visual_cmp}
\end{figure*}

\begin{figure*}[!t]
\centering
\includegraphics[width=0.99\linewidth]{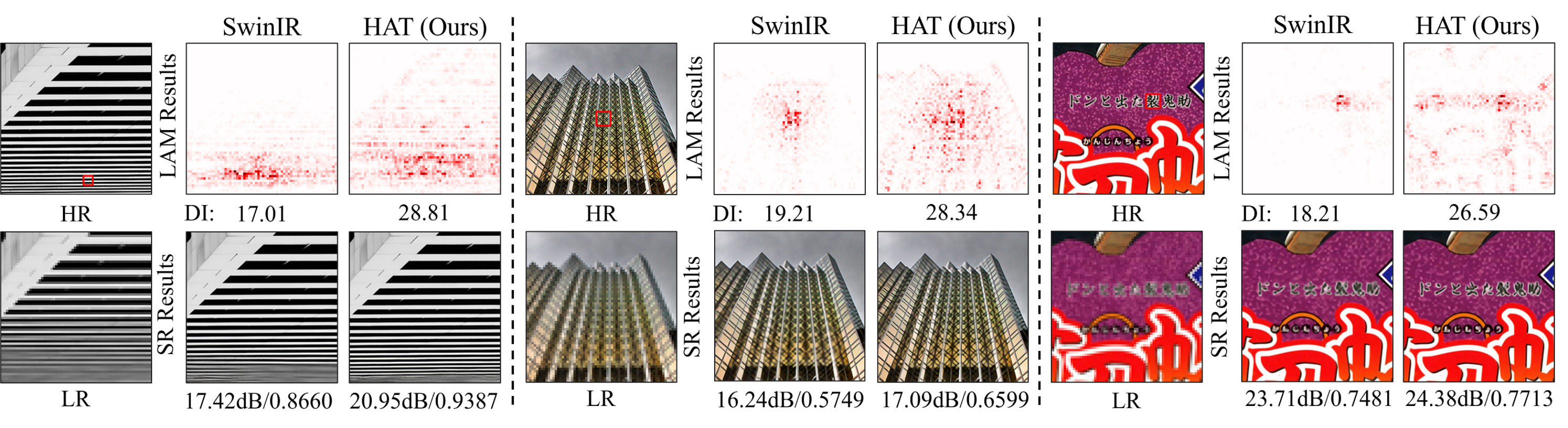}
\vspace{-0.3cm}
\caption{Comparison of LAM results between SwinIR and HAT.}
\label{more_lam}
\vspace{-0.5cm}
\end{figure*}

For image denoising and JPEG compression artifacts reduction, we directly use the combination of DIV2K, Flickr2K, BSD500~\cite{bsd100} and WED images~\cite{wed} datasets to train the models\footnote{ImageNet pre-training has limited impact on the performance of these two tasks. Similar conclusions are also mentioned in~\cite{edt}.}, following~\cite{rdn_pami,drunet,swinir}. The network is the same as the basic version for classic image super-resolution without up-sampling. Noisy images are generated by adding additive white Gaussian noises with noise level $\sigma$ and compressed images are obtained by the MATLAB JPEG encoder with JPEG level $q$. To speed up the training, we first train models with the batch size of 32 and the patch size of $64\times 64$ for 800 iterations. We then proceed to fine-tune models with the batch size of 8 and the patch size of $128\times 128$ for 500 iterations.

\subsection{Classic Image Super-Resolution}
\noindent
\textbf{Quantitative results.}
Tab.~\ref{quantitative results} shows the quantitative comparison of our approach with the state-of-the-art methods: EDSR~\cite{edsr}, RCAN~\cite{rcan}, SAN~\cite{san}, IGNN \cite{ignn}, HAN~\cite{han}, RCAN-it~\cite{rcanit}, GRL~\cite{grl} as well as approaches using ImageNet pre-training, \textit{i.e.}, IPT~\cite{ipt} and EDT~\cite{edt}. We can see that our method significantly outperforms the other methods on all five benchmark datasets. Concretely, with the same depth and width, HAT surpasses SwinIR by 0.48dB$\sim$0.64dB on Urban100 and 0.34dB$\sim$0.45dB on Manga109. When compared with the approaches using pre-training, {HAT}$^\dagger$ also has large performance gains of more than 0.5dB against EDT on Urban100 for all three scales. Besides, HAT equipped with pre-training outperforms SwinIR by a huge margin of up to 1dB on Urban100 for $\times$2 SR. Moreover, the large model HAT-L can even bring further improvement and greatly expands the performance upper bound of this task. HAT-S with fewer parameters and similar computation can also significantly outperforms the state-of-the-art method SwinIR. (Detailed computational complexity comparison can be found in Sec. \ref{complexity}.) Note that the performance gaps are much larger on Urban100, as it contains more structured and self-repeated patterns that can provide more useful pixels for reconstruction when the utilized range of information is enlarged. All these results show the effectiveness of our method. 

\begin{table*}[!t]
\center
\begin{center}
\caption{Quantitative comparison with state-of-the-art methods for \textbf{\underline{real-world image super-resolution}} on benchmark datasets. The best and second best results are marked in \textbf{bold} and \underline{underline}.}
\vspace{-0.1cm}
\label{realsr_quantitative_results}
\renewcommand{\arraystretch}{1.1}
\begin{tabular}{|l|c|c|c|c|c|c|c|c|}
\hline
\multirow{2}{*}{Method} & \multicolumn{2}{c|}{RealSR-cano~\cite{realsrset}} &  \multicolumn{2}{c|}{RealSR-Nikon~\cite{realsrset}} &  \multicolumn{2}{c|}{AIM2019-val~\cite{aim_val}} &  \multicolumn{2}{c|}{DIV2K-SysReal}
\\ 
\cline{2-9}
& PSNR ($\uparrow$) & LPIPS ($\downarrow$) & PSNR ($\uparrow$) & LPIPS ($\downarrow$) & PSNR ($\uparrow$) & LPIPS ($\downarrow$) & PSNR ($\uparrow$) & LPIPS ($\downarrow$)
\\ 
\hline
\hline
ESRGAN~\cite{esrgan}
& \textbf{27.67}
& 0.412
& \textbf{27.46}
& 0.425
& 23.16
& 0.550
& 23.48
& 0.627
\\
DASR~\cite{dasr}
& \underline{27.40}
& 0.393
& 26.35
& 0.401
& 23.76
& 0.421
& 23.78
& 0.473
\\
\hline
BSRGAN~\cite{bsrgan}
& 26.91
& 0.371
& 25.56
& 0.391
& \textbf{24.20}
& 0.400
& \underline{23.83}
& 0.469
\\
SwinIR~\cite{swinir}
& 26.64
& \underline{0.357}
& 25.76
& \underline{0.364}
& 23.89
& 0.387
& 23.31
& 0.449
\\
\textbf{HAT-1} (ours)
& 27.17
& 0.360
& \underline{26.52}
& 0.376
& 24.09
& \underline{0.380}
& 23.62
& \underline{0.439}
\\
\hline
Real-ESRGAN~\cite{realesrgan}
& 26.14
& 0.378
& 25.49
& 0.388
& 23.89
& 0.396
& 23.58
& 0.446
\\
\textbf{HAT-2} (ours)
& 26.68
& \textbf{0.342}
& 25.85
& \textbf{0.358}
& \underline{24.19}
& \textbf{0.370}
& \textbf{23.98}
& \textbf{0.423}
\\
\hline
\end{tabular}
\end{center}
\end{table*}

\begin{figure*}[!t]
\begin{center}
\includegraphics[width=1.0\textwidth]{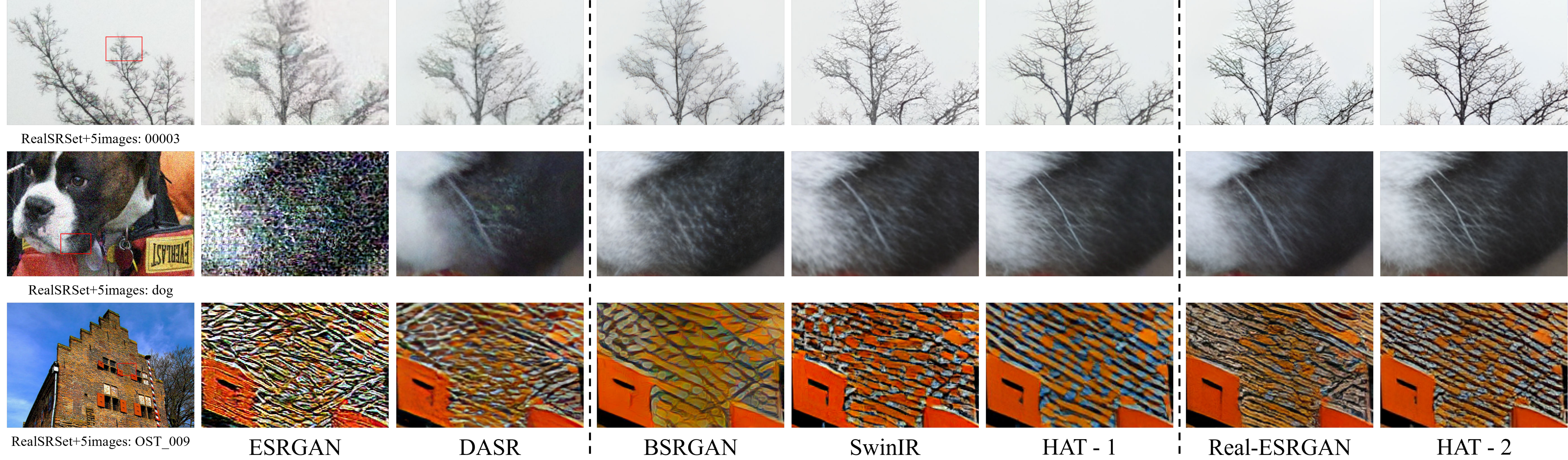}
\end{center}
\vspace{-0.3cm}
\caption{Real-world image super-resolution results on SR $\times4$. HAT-1 uses BSRGAN degradation model and HAT-2 adopts Real-ESRGAN degradation model.}
\vspace{-0.3cm}
\label{realsr}
\end{figure*}

\noindent
\textbf{Visual comparison.}
We provide the visual comparison of different approaches. As shown in Fig.~\ref{visual_cmp}, HAT successfully recovers the clear lattice content for the images ``img\underline{~}002'', ``img\underline{~}011'', ``img\underline{~}030'', ``img\underline{~}044'' and ``img\underline{~}073'' in the Urban100 dataset. In contrast, the other approaches cannot restore correct textures or suffer from severe blurry effects. We can also observe similar behaviors on ``PrayerHaNemurenai'' in the Manga109 dataset. When recovering the characters in the image, HAT obtains much clearer textures than the other methods. The visual results also demonstrate the superiority of our approach 
on classic image super-resolution.

\noindent
\textbf{LAM comparison.}
We provide more visual results with LAM to compare the state-of-the method SwinIR and our HAT. As shown in Fig.~\ref{more_lam}, the utilized pixels for reconstruction of HAT expands to the almost full image, while that of SwinIR only gathers in a limited range. For the quantitative metric, HAT also obtains a much higher DI value than SwinIR. These results demonstrate that our method can activate more pixels to reconstruct the low-resolution input image. As a result, SR results generated by our method have higher PSNR/SSIM and better visual quality. We can observe that HAT restores much clearer textures and edges than SwinIR.

\subsection{Real-world Image Super-Resolution}
\noindent
\textbf{Quantitative results.}
Table \ref{realsr_quantitative_results} presents a quantitative comparison of our method with state-of-the-art approaches: ESRGAN~\cite{esrgan}, BSRGAN~\cite{bsrgan}, Real-ESRGAN~\cite{realesrgan}, DASR~\cite{dasr}, and SwinIR~\cite{swinir}. All models are GAN-based, and our HAT models are trained using the BSRGAN degradation model \cite{bsrgan} (i.e., HAT-1) and Real-ESRGAN degradation model \cite{realesrgan} (i.e., HAT-2). For compared methods, we utilize the officially released models and evaluate them on four datasets. 
Real-SR-cano~\cite{realsrset} and Real-Nikon~\cite{realsrset} feature real-world data pairs from specific cameras. AIM2019-val, from the AIM2019 challenge~\cite{aim_val}, is a synthetic dataset with realistic, unknown degradations\footnote{We use paired data of the validation set from this challenge. The competition officials do not release the degradation model.}. Additionally, we construct a synthetic dataset using 100 DIV2K\underline{~}valid~\cite{div2k} images with a high-order degradation model~\cite{realesrgan}, following DASR~\cite{dasr}. We use PSNR and LPIPS~\cite{lpips} as metrics, with PSNR measuring fidelity and LPIPS assessing perceptual quality.
Since all of the methods are GAN-based, the PSNR performance of different models on various datasets is not consistent. Nevertheless, similar PSNR values suggest comparable fidelity among the methods. Notably, HAT-1 achieves the best balance between PSNR and LPIPS among the three methods. HAT-2 obtains the best performance on all four datasets, indicating that it generates the results with the best perceptual quality. 

\begin{table*}[!t]\scriptsize
\center
\begin{center}
\caption{Quantitative comparison (average PSNR) with state-of-the-art methods for \textbf{\underline{grayscale image denoising}} on benchmark datasets. The best nad second best results are marked in \textbf{bold} and \underline{underline}, respectively.}
\vspace{-0.2cm}
\label{tab:denoising_grayscale_results}
\renewcommand{\arraystretch}{1.1}
\resizebox{\textwidth}{21mm}{
\begin{tabular}
{|c|c|c|c|c|c|c|c|c|c|c|c|c|c|c|}
\hline
~Dataset~ & ~$\sigma$~ & 
\makecell{DnCNN\\\cite{dncnn}} & 
\makecell{IRCNN\\\cite{ircnn}} &
\makecell{FFDNet\\\cite{ffdnet}} &
\makecell{NLRN\\\cite{nlrn}} &
\makecell{MWCNN\\\cite{mwcnn}} & 
\makecell{DRUNet\\\cite{drunet}} &
\makecell{SwinIR\\\cite{swinir}} &
\makecell{Restormer\\\cite{restormer}} &
\makecell{SCUNet\\\cite{scunet}} &
\makecell{GRL\\\cite{grl}} &
\makecell{\textbf{HAT}\\(ours)} 
\\
\hline
\hline
\multirow{3}{*}{\makecell{Set12\\\cite{dncnn}}} & 15
& 32.86
& 32.76 
& 32.75
& 33.16 
& 33.15
& 33.25
& 33.36
& 33.42
& 33.43
& \underline{33.47}
& \textbf{33.49}
\\
& 25
& 30.44
& 30.37
& 30.43
& 30.80
& 30.79
& 30.94
& 31.01
& 31.08
& 31.09
& \underline{31.12}
& \textbf{31.13}
\\
& 50
& 27.18
& 27.12
& 27.32 
& 27.64
& 27.74
& 27.90
& {27.91}
& 28.00
& \underline{28.04}
& 28.03
& \textbf{28.07}
\\
\hline
\multirow{3}{*}{\makecell{BSD68\\\cite{bsd100}}} & 15
& 31.73
& 31.63
& 31.63
& 31.88
& 31.86
& 31.91
& 31.97
& 31.96
& \underline{31.99}
& \textbf{32.00}
& \underline{31.99}
\\
& 25
& 29.23
& 29.15
& 29.19
& 29.41
& 29.41
& 29.48
& 29.50
& 29.52
& \textbf{29.55}
& \underline{29.54}
& 29.52
\\
& 50
& 26.23
& 26.19
& 26.29
& 26.47
& 26.53
& 26.59
& 26.58
& \underline{26.62}
& \textbf{26.67}
& 26.60
& 26.60
\\
\hline
\multirow{3}{*}{\makecell{\hspace{-0.15cm}Urban100\\\cite{urban100}}} & 15
& 32.64
& 32.46
& 32.40
& 33.45
& 33.17
& 33.44
& {33.70}
& 33.79
& 33.88
& \textbf{34.09}
& \underline{33.99}
\\
& 25
& 29.95
& 29.80
& 29.90
& 30.94
& 30.66
& 31.11
& 31.30
& 31.46
& 31.58
& \textbf{31.80}
& \underline{31.67}
\\
& 50
& 26.26
& 26.22
& 26.50
& 27.49
& 27.42
& 27.96
& 27.98
& 28.29
& 28.56
& \underline{28.59}
& \textbf{28.62}
\\
\hline             
\end{tabular}
}
\end{center}
\vspace{-0.2cm}
\end{table*}

\begin{table*}[!t]
\center
\begin{center}
\caption{Quantitative comparison (average PSNR) with state-of-the-art methods for \textbf{\underline{color image denoising}} on benchmark datasets. The best, second best results are marked in \textbf{bold} and \underline{underline}, respectively.}
\vspace{-0.2cm}
\label{tab:denoising_color_results}
\renewcommand{\arraystretch}{1.1}
\resizebox{\textwidth}{26mm}{
\begin{tabular}{|c|c|c|c|c|c|c|c|c|c|c|c|c|c|c|c}
\hline
~~~~~Dataset~~~~~ & ~$\sigma$~ & 
\makecell{DnCNN\\\cite{dncnn}} &
\makecell{IRCNN\\\cite{ircnn}} &
\makecell{FFDNet\\\cite{ffdnet}} &
\makecell{RNAN\\\cite{rnan}} &
\makecell{RDN\\\cite{rdn}} &
\makecell{IPT\\\cite{ipt}} &
\makecell{DRUNet\\\cite{drunet}} &
\makecell{SwinIR\\\cite{swinir}} &
\makecell{Restormer\\\cite{restormer}} &
\makecell{SCUNet\\\cite{scunet}} &
\makecell{\textbf{HAT}\\(ours)}
\\
\hline
\hline
\multirow{3}{*}{\makecell{CBSD68\\\cite{bsd100}}} & 15
& 33.90
& 33.86
& 33.87
& -
& -
& -
& {34.30}
& \textbf{34.42}
& \underline{34.40}
& \underline{34.40}
& \textbf{34.42}
\\
& 25
& 31.24 
& 31.16
& 31.21
& -
& -
& -
& {31.69}
& \underline{31.78}
& \textbf{31.79}
& \textbf{31.79}
& \textbf{31.79}
\\
& 50
& 27.95
& 27.86
& 27.96
& 28.27
& 28.31
& 28.39
& 28.51
& {28.56}
& \underline{28.60}
& \textbf{28.61}
& 28.58
\\
\hline
\multirow{3}{*}{\makecell{Kodak24\\\cite{kodak}}} & 15
& 34.60 
& 34.69
& 34.63
& -
& -
& -
& 35.31
& 35.34
& \underline{35.47}
& 35.34
& \textbf{35.50}
\\
& 25
& 32.14
& 32.18
& 32.13
& -
& -
& -
& 32.89
& 32.89
& \underline{33.04}
& 32.92
& \textbf{33.08}
\\
& 50
& 28.95
& 28.93
& 28.98
& 29.58
& 29.66
& 29.64
& {29.86}
& 29.79
& \underline{30.01}
& 29.87
& \textbf{30.02}
\\
\hline
\multirow{3}{*}{\makecell{McMaster\\\cite{mcmaster}}} & 15
& 33.45
& 34.58
& 34.66
& -
& -
& -
& 35.40
& \underline{35.61}
& \underline{35.61}
& 35.60
& \textbf{35.64}
\\
& 25
& 31.52
& 32.18
& 32.35
& -
& -
& -
& 33.14
& {33.20}
& \underline{33.34}
& \underline{33.34}
& \textbf{33.37}
\\
& 50
& 28.62 
& 28.91
& 29.18
& 29.72
& -
& 29.98
& 30.08
& {30.22}
& \textbf{30.30}
& \underline{30.29}
& 30.26
\\
\hline
\multirow{3}{*}{\makecell{Urban100\\\cite{urban100}}} & 15
& 32.98
& 33.78
& 33.83
& -
& -
& -
& {34.81}
& 35.13
& 35.13
& \underline{35.18}
& \textbf{35.37}
\\
& 25
& 30.81
& 31.20
& 31.40
& -
& -
& -
& 32.60
& 32.90
& 32.96
& \underline{33.03}
& \textbf{33.14}
\\
& 50
& 27.59
& 27.70
& 28.05
& 29.08
& 29.38
& 29.71 
& 29.61
& {29.82}
& 30.02
& \underline{30.14}
& \textbf{30.20}
\\
\hline             
\end{tabular}
}
\end{center}
\vspace{-0.15cm}
\end{table*}

\begin{table*}[!t]
\center
\begin{center}
\caption{Quantitative comparison (average PSNR/SSIM) with state-of-the-art methods for \textbf{\underline{JPEG compression artifacts reduction}} on benchmark datasets. The best and second best results are marked in \textbf{bold} and \underline{underline}, respectively.}
\vspace{-0.1cm}
\label{tab:dejpeg_results}
\renewcommand{\arraystretch}{1.2}
\begin{tabular}
{|p{1.1cm}<{\centering}|p{0.25cm}<{\centering}|p{1.8cm}<{\centering}|p{1.8cm}<{\centering}|p{1.8cm}<{\centering}|p{1.8cm}<{\centering}|p{1.8cm}<{\centering}|p{1.8cm}<{\centering}|p{1.8cm}<{\centering}|p{1.8cm}<{\centering}}
\hline
Dataset & $q$ 
& ARCNN~\cite{arcnn} 
& DnCNN~\cite{dncnn} 
& RNAN~\cite{rnan}
& FBCNN~\cite{fbcnn}
& DRUNet~\cite{drunet}
& SwinIR~\cite{swinir}
& \textbf{HAT} (ours)\\
\hline
\hline
\multirow{4}{*}{\makecell{Classic5\\\cite{classic5}}} & 10
& 29.03/0.7929
& 29.40/0.8026
& 29.96/0.8178
& 30.12/0.8223
& \underline{30.16}/0.8234
& \textbf{30.27}/\textbf{0.8249}
& \textbf{30.27}/\underline{0.8246}
\\
& 20
& 31.15/0.8517
& 31.63/0.8610
& 32.11/0.8693
& 32.31/0.8724
& 32.39/0.8734
& \underline{32.52}/\underline{0.8748}
& \textbf{32.54}/\textbf{0.8749}
\\
& 30
& 32.51/0.8806
& 32.91/0.8861
& 33.38/0.8924
& 33.54/0.8943
& 33.59/0.8949
& \textbf{33.73}/\textbf{0.8961}
& \underline{33.72}/\underline{0.8958}
\\
& 40
& 33.32/0.8953
& 33.77/0.9003
& 34.27/0.9061
& 34.35/0.9070
& 34.41/0.9075
& \underline{34.52}/\underline{0.9082}
& \textbf{34.58}/\textbf{0.9086}
\\
\hline
\multirow{4}{*}{\makecell{LIVE1\\\cite{live1}}} & 10
& 28.96/0.8076
& 29.19/0.8123
& 29.63/0.8239
& 29.75/0.8268
& 29.79/0.8278
& \textbf{29.86}/\textbf{0.8287}
& \underline{29.84}/\underline{0.8283}
\\
& 20
& 31.29/0.8733
& 31.59/0.8802
& 32.03/0.8877
& 32.13/0.8893
& 32.17/0.8899
& \underline{32.25}/\textbf{0.8909}
& \textbf{32.26}/\underline{0.8907}
\\
& 30
& 32.67/0.9043
& 32.98/0.9090
& 33.45/0.9149
& 33.54/0.9161
& 33.59/0.9166
& \textbf{33.69}/\textbf{0.9174}
& \underline{33.66}/\underline{0.9170}
\\
& 40
& 33.63/0.9198
& 33.96/0.9247
& 34.47/0.9299
& 34.53/0.9307
& 34.58/\underline{0.9312}
& \underline{34.67}/\textbf{0.9317}
& \textbf{34.69}/\textbf{0.9317}
\\
\hline
\multirow{4}{*}{\makecell{Urban100\\\cite{urban100}}} & 10
& -
& 28.54/0.8487
& 29.76/0.8723
& 30.15/0.8795
& 30.05/0.8772
& \underline{30.55}/\underline{0.8842}
& \textbf{30.60}/\textbf{0.8845}
\\
& 20
& -
& 31.01/0.9022
& 32.33/0.9179
& 32.66/0.9219
& 32.66/0.9216
& \underline{33.12}/\underline{0.9254}
& \textbf{33.22}/\textbf{0.9260}
\\
& 30
& -
& 32.47/0.9248
& 33.83/0.9365
& 34.09/0.9392
& \underline{34.13}/0.9392
& \textbf{34.58}/\underline{0.9417}
& \textbf{34.58}/\textbf{0.9419}
\\
& 40
& -
& 33.49/0.9376
& 34.95/0.9476
& 35.08/0.9490
& 35.11/0.9491
& \underline{35.50}/\underline{0.9509}
& \textbf{35.68}/\textbf{0.9517}
\\
\hline
\end{tabular}
\end{center}
\vspace{-0.35cm}
\end{table*}

\begin{figure*}[!t]
\begin{center}
\includegraphics[width=1.0\textwidth]{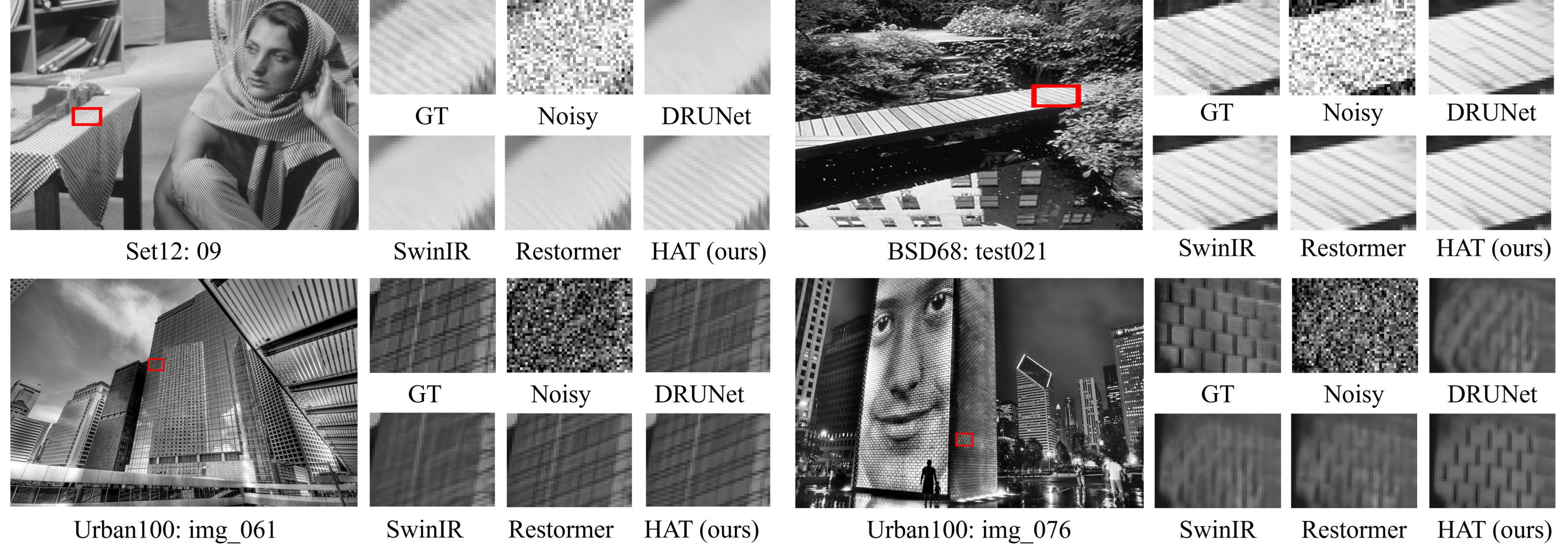}
\end{center}
\vspace{-0.5cm}
\caption{Grayscale image denoising results with noise level $\sigma=50$.}
\vspace{-0.3cm}
\label{gray_denoising}
\end{figure*}

\begin{figure*}[!t]
\begin{center}
\includegraphics[width=1.0\textwidth]{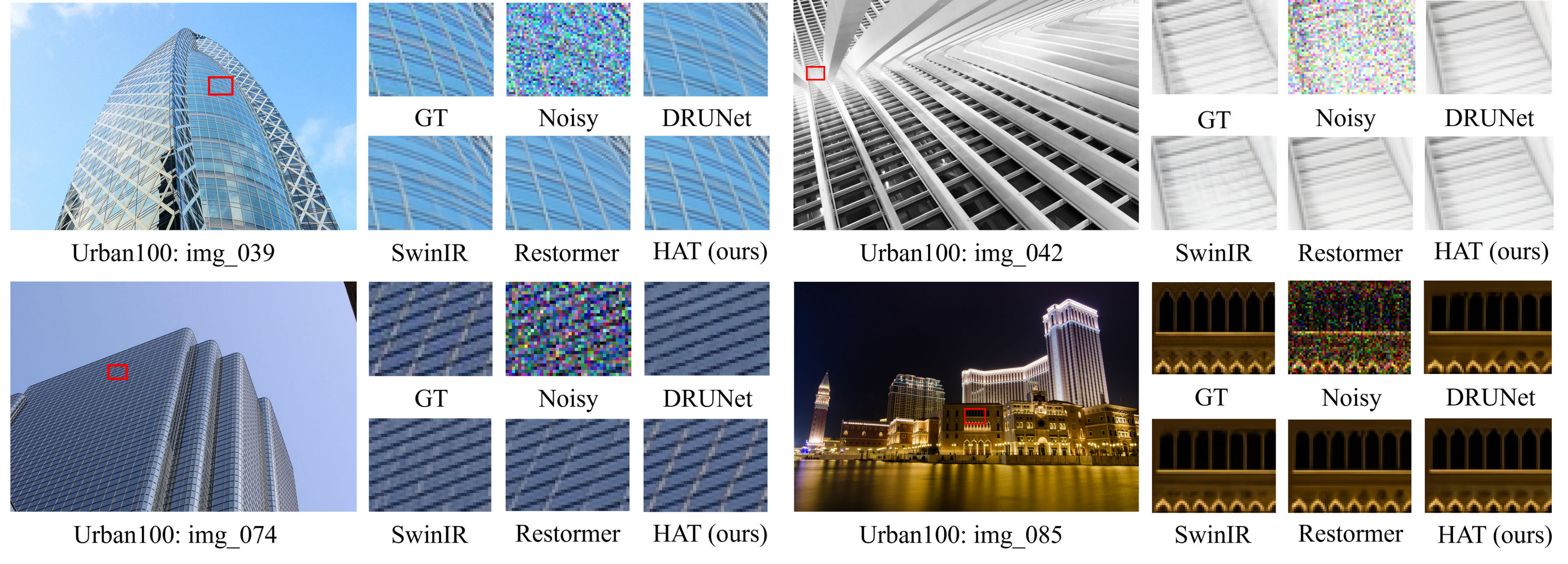}
\end{center}
\vspace{-0.5cm}
\caption{Color image denoising results with noise level $\sigma=50$.}
\vspace{-0.3cm}
\label{color_denoising}
\end{figure*}

\begin{figure*}[!t]
\begin{center}
\includegraphics[width=1.0\textwidth]{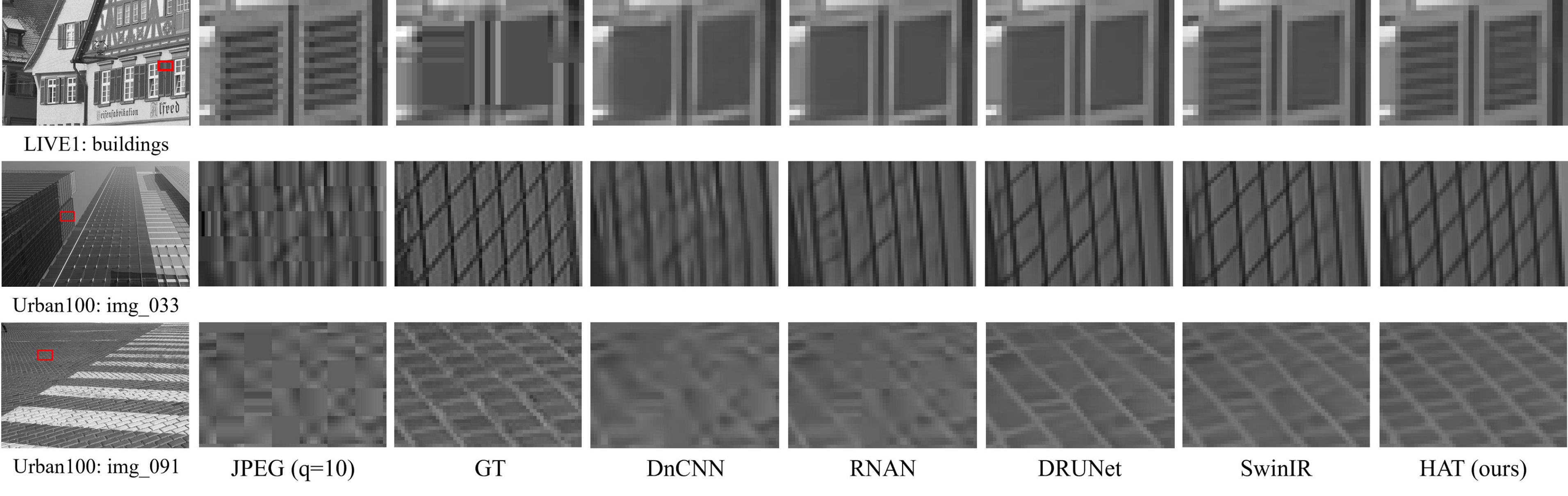}
\end{center}
\vspace{-0.5cm}
\caption{Image compression artifacts reduction results with JPEG quality $q=10$.}
\vspace{-0.5cm}
\label{dejpeg}
\end{figure*}

\begin{figure}[!t]
\begin{center}
\includegraphics[width=1\linewidth]{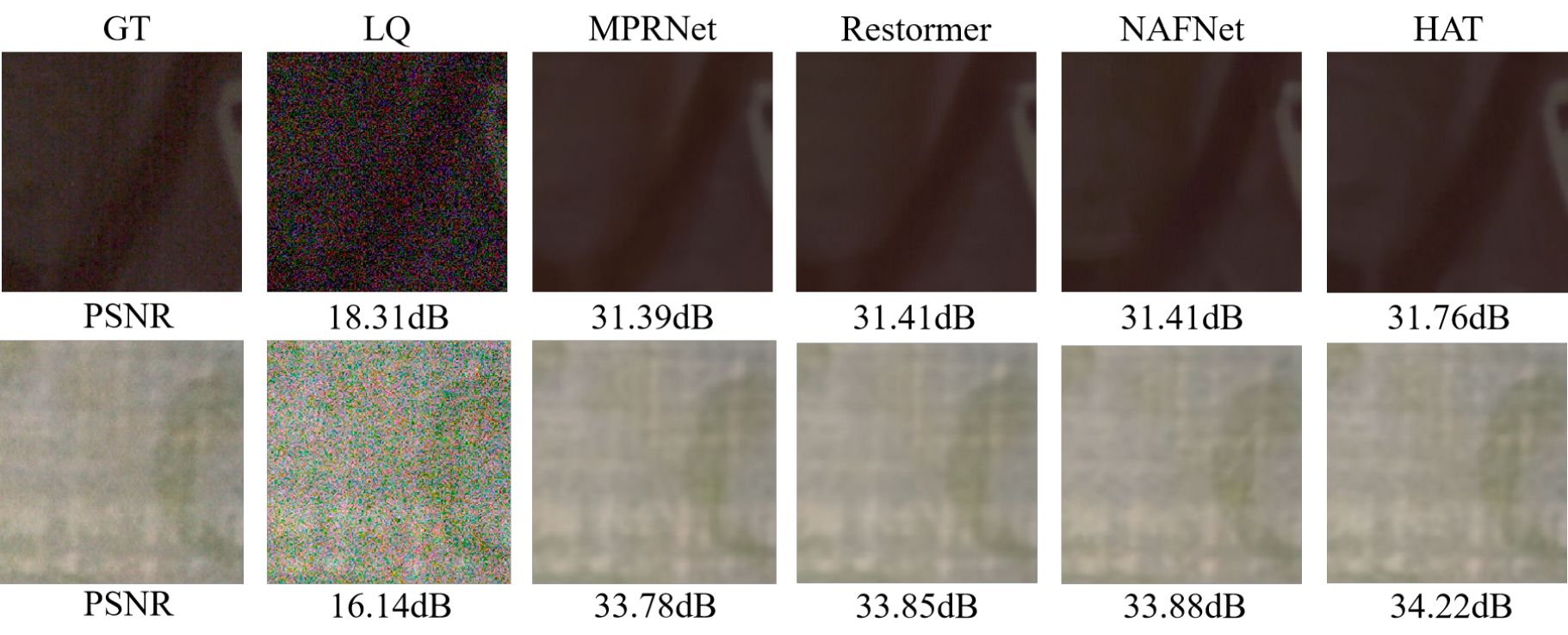}
\end{center}
\vspace{-0.5cm}
\caption{Real-world image denoising results on the SIDD dataset~\cite{sidd}.}
\vspace{-1cm}
\label{tab:sidd_denoising}
\end{figure}

\noindent
\textbf{Visual comparison.} 
We show the visual results from different methods on real-world low-resolution images in Fig.~\ref{realsr}. We adopt the RealSRSet+5images~\cite{bsrgan} as the test set, which is commonly used for evaluating real-world SR models. As different degradation models from BSRGAN~\cite{bsrgan} and Real-ESRGAN~\cite{realsrset} may produce varied visual properties, we present results for both models, denoted as HAT-1 and HAT-2.
In the first and second rows of visual comparisons, our HAT produces much clearer branches and whiskers than other methods. In the third row, results from different degradation models exhibit significant variation. BSRGAN-based models show notable differences in color and texture. The details of BSRGAN results are relatively smooth. SwinIR generates clear textures but with obvious color deviations. In contrast, HAT achieves a relatively good balance. It handles the details well, and its larger receptive field also enhances the processing accuracy of low-frequency information. Using the degradation model from Real-ESRGAN, we can see that the result of HAT appear neater and more brick-like. Overall, our method produces visually appealing results with sharp, clear edges, demonstrating its potential for real-world applications.

\subsection{Image Denoising}
\subsubsection{Grayscale Image Denoising}
Tab.~\ref{tab:denoising_grayscale_results} shows the quantitative comparison on grayscale image denoising of our approach with the state-of-the-art method: 
DnCNN~\cite{dncnn}, IRCNN~\cite{ircnn}, FFDNet~\cite{ffdnet}, NLRN~\cite{nlrn}, MWCNN~\cite{mwcnn}, DRUNet~\cite{drunet}, SwinIR~\cite{swinir}, Restormer~\cite{restormer} , GRL~\cite{grl} and SCUNet~\cite{scunet}. 
Compared to SwinIR~\cite{swinir}, HAT achieves better performance on all datasets with multiple noise levels. On Urban100, HAT achieves the largest performance gain by up to 0.64dB for $\sigma$ 50.
Compared to current state-of-the-art methods Restormer~\cite{restormer} and SCUNet~\cite{scunet}, HAT can still outperform the former and obtains comparable performance with the latter. 
We provide the visual results of different methods in Fig.~\ref{gray_denoising}. For ``09'' in Set12, our HAT restores clear lines while other approaches suffers from severe blurs. For ``test021'' in BSD68 and ``img\underline{~}061'' in Urban100, HAT reconstructs much sharper edges than other methods. For ``img\underline{~}076'' in Urban100, the results produced by HAT have the clearest textures. Overall, HAT obtains the best visual quality among all methods. 

\subsubsection{Color Image Denoising}
Tab.~\ref{tab:denoising_color_results} shows the quantitative comparison results on color image denoising of our approach with the state-of-the-art methods: DnCNN~\cite{dncnn}, IRCNN~\cite{ircnn}, FFDNet~\cite{ffdnet}, RNAN~\cite{rnan}, 
RDN~\cite{rdn_pami}, IPT~\cite{ipt}, DRUNet~\cite{drunet}, SwinIR~\cite{swinir}, Restormer~\cite{restormer} and SCUNet~\cite{scunet}. We can observe that HAT achieves the best performance on almost all four benchmark datasets. Specifically, HAT outperforms SwinIR from 0.24dB to 0.38dB and surpasses SCUNet by a large margin of 0.19dB on Urban100 with $\sigma=15$. 

We present the visual results of different methods in Fig.~\ref{color_denoising}. For images ``img\underline{~}039'', ``img\underline{~}042'' and ``img\underline{~}074'' on Urban100, HAT reconstructs complete and clear edges, whereas other methods cannot produce complete lines or suffer from severe blurring effects. For the image ``img\underline{~}085'', our method successfully restores the correct shape and clear texture, while other methods all fail. All these results demonstrate the superiority of the proposed HAT on image denoising.

\subsubsection{Real-world Image Denoising}
We also evaluate our proposed method for real-world image denoising based on the SIDD dataset~\cite{sidd}. The compared methods include MPRNet~\cite{mprnet}, UFormer~\cite{uformer}, MAXIM~\cite{maxim}, HINet~\cite{hinet}, Restormer~\cite{restormer}, and NAFNet~\cite{nafnet}. The quantitative and qualitative results are shown in Tab.~\ref{tab:sidd_denoising} and Fig.~\ref{fig:sidd_denoising}. Our method achieves the highest PSNR/SSIM scores and clearest results, demonstrating superior performance in real-world scenarios. 

\begin{table*}[!t]
\center
\begin{center}
\caption{Real-world image denoising results on the SIDD dataset~\cite{sidd}.}
\vspace{-0.5cm}
\label{fig:sidd_denoising}
\setlength{\tabcolsep}{3.4mm}{
\begin{tabular}{c|ccccccc} 
\hline 
 Method & MPRNet~\cite{mprnet} & UFormer~\cite{uformer} & MAXIM~\cite{maxim} & HINet~\cite{hinet} & Restormer~\cite{restormer} & NAFNet~\cite{nafnet} & \textbf{HAT} (ours) \\
\hline 
PSNR(dB)/SSIM & 39.71/0.958 & 39.89/0.960 & 39.96/0.960 & 39.99/0.958 & 40.02/0.960 & 40.30/0.962 & \textbf{40.50/0.964}\\
\hline 
\end{tabular}
}
\end{center}
\vspace{-0.25cm}
\end{table*}

\begin{table*}[!t]
\center
\begin{center}
\caption{Image deblurring results on the Gopro dataset~\cite{Gopro}.}
\vspace{-0.5cm}
\label{gopro_deblurring}
\setlength{\tabcolsep}{2.3mm}{
\begin{tabular}{c|cccccccc} 
\hline 
 Method & MPRNet~\cite{mprnet} & HINet~\cite{hinet} & MAXIM~\cite{maxim} & Restormer~\cite{restormer} & UFormer~\cite{uformer} & NAFNet~\cite{nafnet} & GRL~\cite{grl} & \textbf{HAT} (ours) \\
\hline 
PSNR(dB)/SSIM & 32.66/0.959 & 32.71/0.959 & 32.86/0.961 & 32.92/0.961 & 32.97/0.967 & 33.69/0.967 & 33.93/\textbf{0.968} & \textbf{33.96/0.968}\\
\hline 
\end{tabular}
}
\end{center}
\vspace{-0.25cm}
\end{table*}

\begin{figure*}[!t]
\begin{center}
\includegraphics[width=1.0\textwidth]{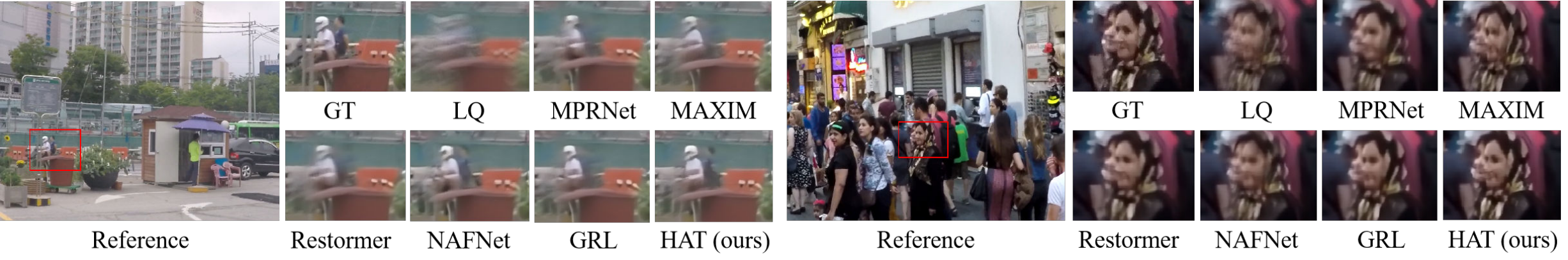}
\end{center}
\vspace{-0.3cm}
\caption{Image deblurring results on the Gopro dataset~\cite{Gopro}.}
\vspace{-0.5cm}
\label{deblur}
\end{figure*}

\subsection{JPEG Compression Artifacts Reduction}
Tab.~\ref{tab:dejpeg_results} shows the quantitative comparison results on JPEG compression artifacts reduction of our approach with the state-of-the-art methods: ARCNN~\cite{arcnn}, DnCNN~\cite{dncnn}, RNAN~\cite{rnan}, 
DRUNet~\cite{drunet}, 
FBCNN~\cite{fbcnn},
SwinIR~\cite{swinir}. On Classic5 and LIVE1, HAT only achieves comparable performance to SwinIR. We consider this is because the demand for the model fitting ability of this task has approached saturation, particularly for low-resolution images. We further provide the performance comparison on Urban100. Then we can see that HAT achieves considerable performance gains over SwinIR, up to 0.18dB for JPEG quality $q=40$. 
This can be attributed to the presence of a large number of regular textures and repeating patterns in the images of the Urban100 dataset. HAT is capable of activating more pixels for restoration. With a larger receptive field, it can restore sharper edges and textures. 

\noindent
We also provide the visual results of different approaches in Fig.~\ref{dejpeg}. For the image ``buildings'' in LIVE1, HAT obtains much clearer textures than other methods. For the self-repeated textures that appear in the images ``img\underline{~}033'' and `img\underline{~}091'' of Urban100, our HAT successfully restores the correct results. All the quantitative and visual results demonstrate the superiority of our method on compression artifacts reduction. 

\subsection{Image Deblurring}
We conduct experiments to further validate the effectiveness of our method on image deblurring. Specifically, we evaluate several representative deblurring methods on the GoPro dataset~\cite{Gopro}. The compared methods include MPRNet~\cite{mprnet}, HINet~\cite{mprnet}, MAXIM~\cite{maxim}, Restormer~\cite{restormer}, Uformer~\cite{uformer}, NAFNet~\cite{nafnet}, and GRL~\cite{grl}.
As shown in Tab.~\ref{gopro_deblurring}, our HAT achieves the best performance among all compared approaches, with a PSNR of 33.96 dB. In addition, as illustrated in Fig.~\ref{deblur}, HAT successfully reconstructs sharp and clear human silhouettes and portraits, recovering visually plausible details from heavily blurred inputs.
These results confirm that our method not only achieves state-of-the-art performance in tasks such as SR, denoising, and compression artifact reduction, but also generalizes effectively to image deblurring. This further validates the superiority of our design.

\section{Conclusion}
In this work, we propose a new Hybrid Attention Transformer, HAT, for image restoration. Our model combines channel attention and self-attention to activate more pixels for high-resolution reconstruction. Besides, we propose an overlapping cross-attention module to enhance the cross-window interaction. Moreover, we introduce a same-task pre-training strategy for image super-resolution. Extensive benchmark and real-world evaluations demonstrate that HAT outperforms the state-of-the-art methods for several image restoration tasks.

\section*{Acknowledgments}
This work was supported in part by National Natural Science Foundation of China (Grant No.62276251, 62272450); in part by Macau Science and Technology Development Fund under 001/2024/SKL, 0022/2022/A1, and 0119/2024/RIB2; in part by Research Committee at University of Macau under MYRG-GRG2023-00058-FST-UMDF; in part by the Guangdong Basic and Applied Basic Research Foundation under Grant 2024A1515012536. 

\bibliographystyle{IEEEtran}
\bibliography{bibtex}

\vspace{-0.5cm}
\begin{IEEEbiography}[{\includegraphics[width=1in,height=1.25in,clip,keepaspectratio]{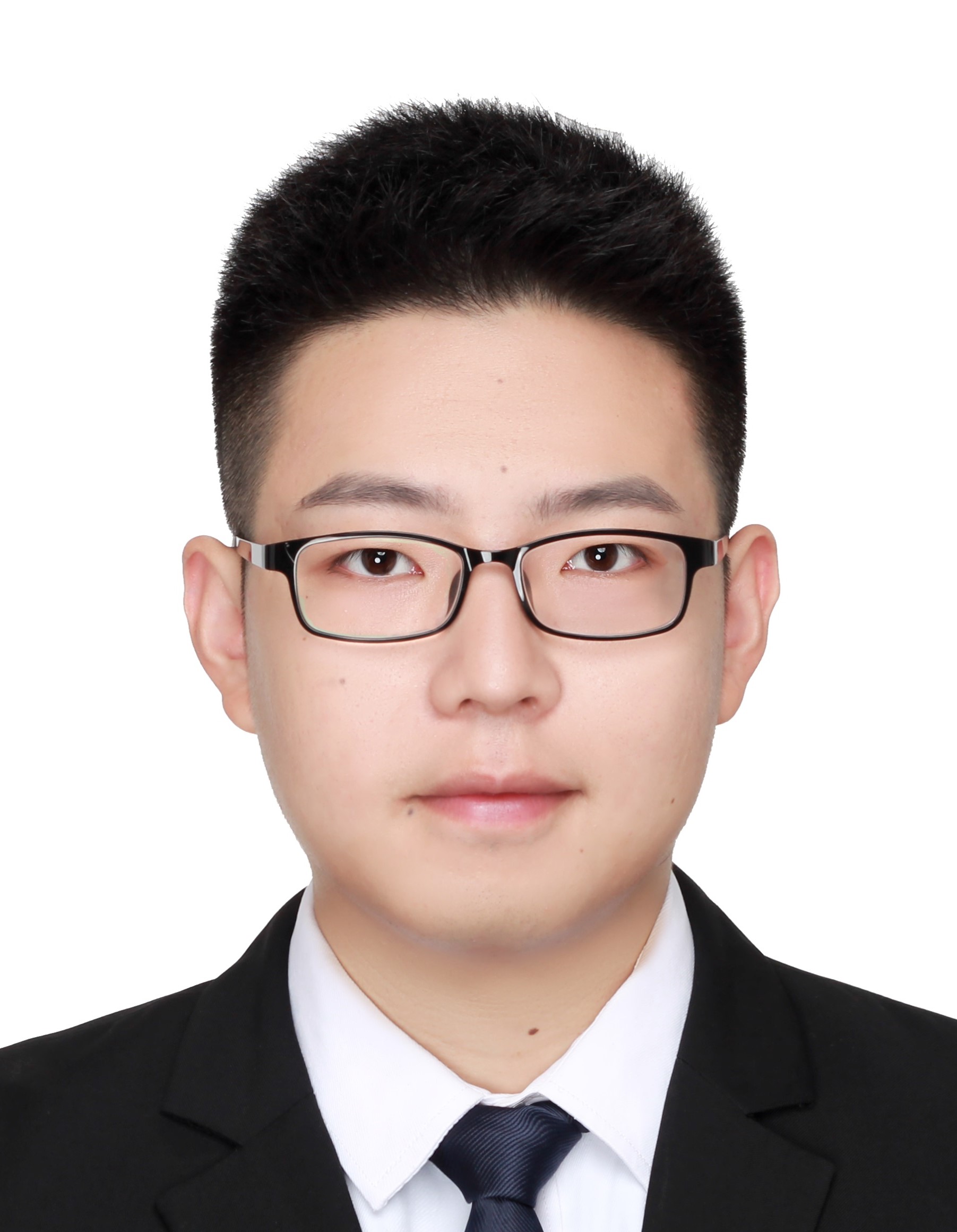}}]{Xiangyu Chen} 
is currently a Research Scientist at the Institute of Artificial Intelligence (TeleAI), China Telecom. He received his Ph.D. at University of Macau in 2025. 
He received his B.E. degree and M.E. degree from Northwestern Polytechnical University, Xi'an, in 2017 and 2020. He worked as research assistant in Multimedia Laboratory, Shenzhen Institute of Advanced Technology, Chinese Academy of Sciences from 2019 to 2021. 
His research interests includes general low-level vision and large multimodal model.
\end{IEEEbiography}

\vspace{-0.5cm}
\begin{IEEEbiography}[{\includegraphics[width=1in,height=1.25in,clip,keepaspectratio]{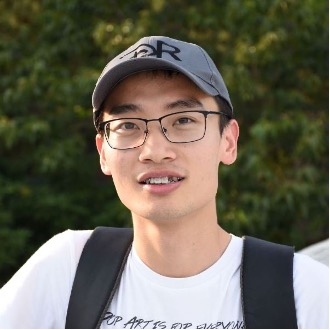}}]{Xintao Wang} 
is currently a senior staff researcher at KwaiVGI, Kuaishou Technology. He received his Ph.D. degree in the Department of Information Engineering, The Chinese University of Hong Kong, in 2020.  He won several champions in international super-resolution challenges, such as NTIRE2019, NTIRE2018, and PIRM2018. His research primarily focus on visual content generation, as well as image and video restoration and enhancement.
\end{IEEEbiography}

\begin{IEEEbiography}[{\includegraphics[width=1in,height=1.25in,clip,keepaspectratio]{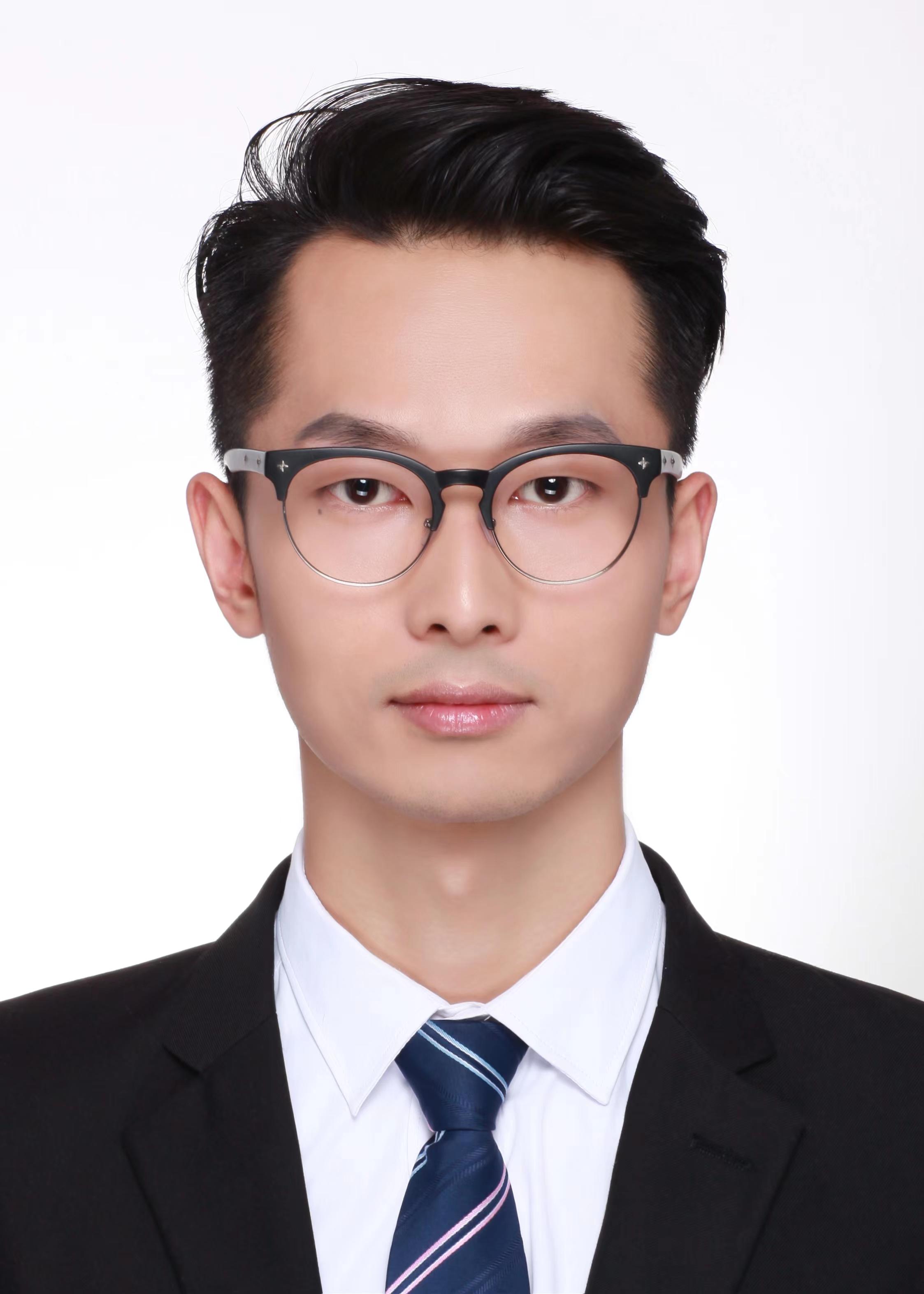}}]{Wenlong Zhang} 
is currently a young researcher at Shanghai AI Laboratory. He receives his Ph.D. degree of HongKong Polytechnic University in 2024. He received his M.S. degree from Beijing Institute of Technology, Beijing, in 2018. He worked as research assistant in Multimedia Laboratory, Shenzhen Institute of Advanced Technology, Chinese Academy of Sciences from 2018 to 2020. His research interests include image super-resolution and restoration.
\end{IEEEbiography}

\begin{IEEEbiography}[{\includegraphics[width=1in,height=1.25in,clip,keepaspectratio]{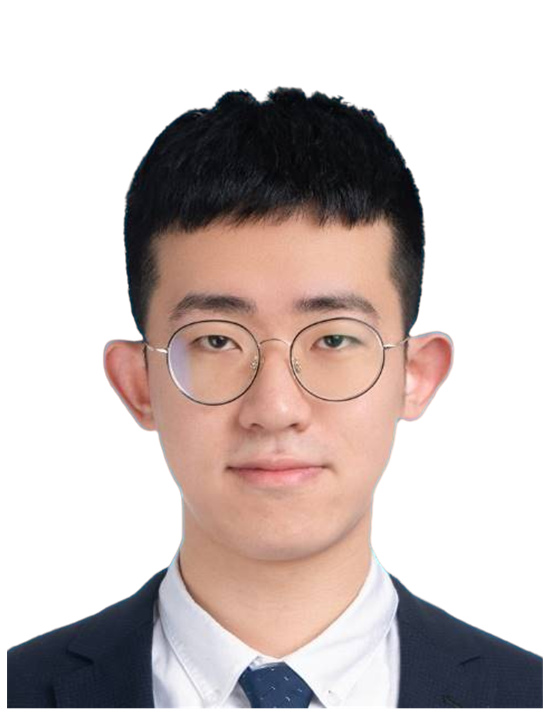}}]{Xiangtao Kong} is currently a Ph.D. student at The Hong Kong Polytechnic University, under the supervision of Prof. Lei Zhang. He received the B.Eng. degree from Shandong University in 2020, and the M.Phil. degree from Shenzhen Institute of Advanced Technology, Chinese Academy of Sciences in 2023, supervised by Prof. Chao Dong. His research interests include image processing and low-level computer vision.
\end{IEEEbiography}

\begin{IEEEbiography}[{\includegraphics[width=1in,height=1.25in,clip,keepaspectratio]{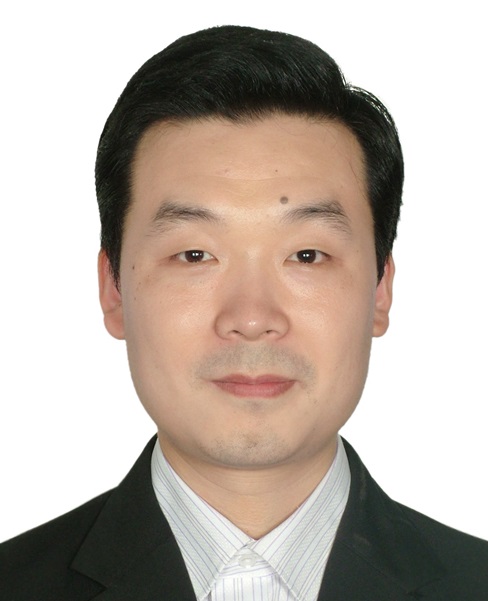}}]{Yu Qiao} (Senior Member, IEEE) 
is a professor with Shanghai AI Laboratory and the Shenzhen Institute of Advanced Technology (SIAT), Chinese Academy of Sciences. He has published more than 600 articles in international journals and conferences, including T-PAMI, IJCV, T-IP, T-SP, CVPR, and ICCV. His research interests include computer vision, deep learning, and bioinformation. He received the First Prize of the Guangdong Technological Invention Award, and the Jiaxi Lv Young Researcher Award from the Chinese Academy of Sciences.
\end{IEEEbiography}

\begin{IEEEbiography}[{\includegraphics[width=1in,height=1.25in,clip,keepaspectratio]{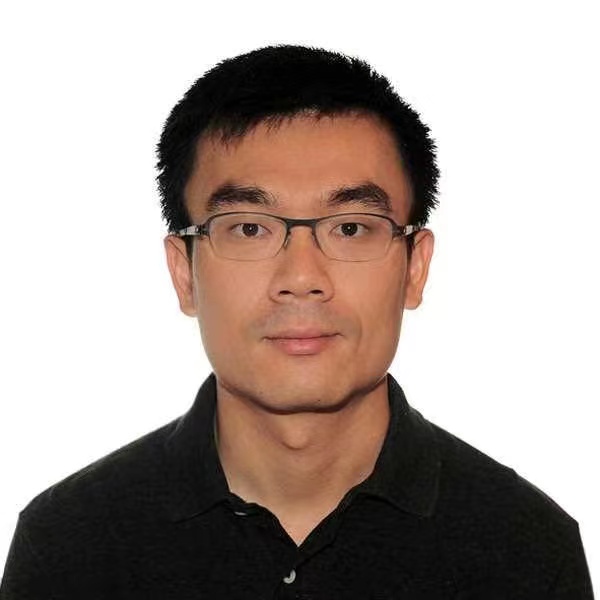}}]{Jiantao Zhou} received the B.Eng. degree from the Department of Electronic Engineering, Dalian University of Technology, in 2002, the M.Phil. degree from the Department of Radio Engineering, Southeast University, in 2005, and the Ph.D. degree from the Department of Electronic and Computer Engineering, Hong Kong University of Science and Technology, in 2009. He held various research positions with the University of Illinois at Urbana-Champaign, Hong Kong University of Science and Technology, and McMaster University. He is now a Professor with the Department of Computer and Information Science, Faculty of Science and Technology, University of Macau. His research interests include multimedia security and forensics, multimedia signal processing, artificial intelligence and big data. 
He is serving as an Associate Editor for the IEEE T-IP, T-MM and T-DSC.
\end{IEEEbiography}

\begin{IEEEbiography}[{\includegraphics[width=1in,height=1.25in,clip,keepaspectratio]{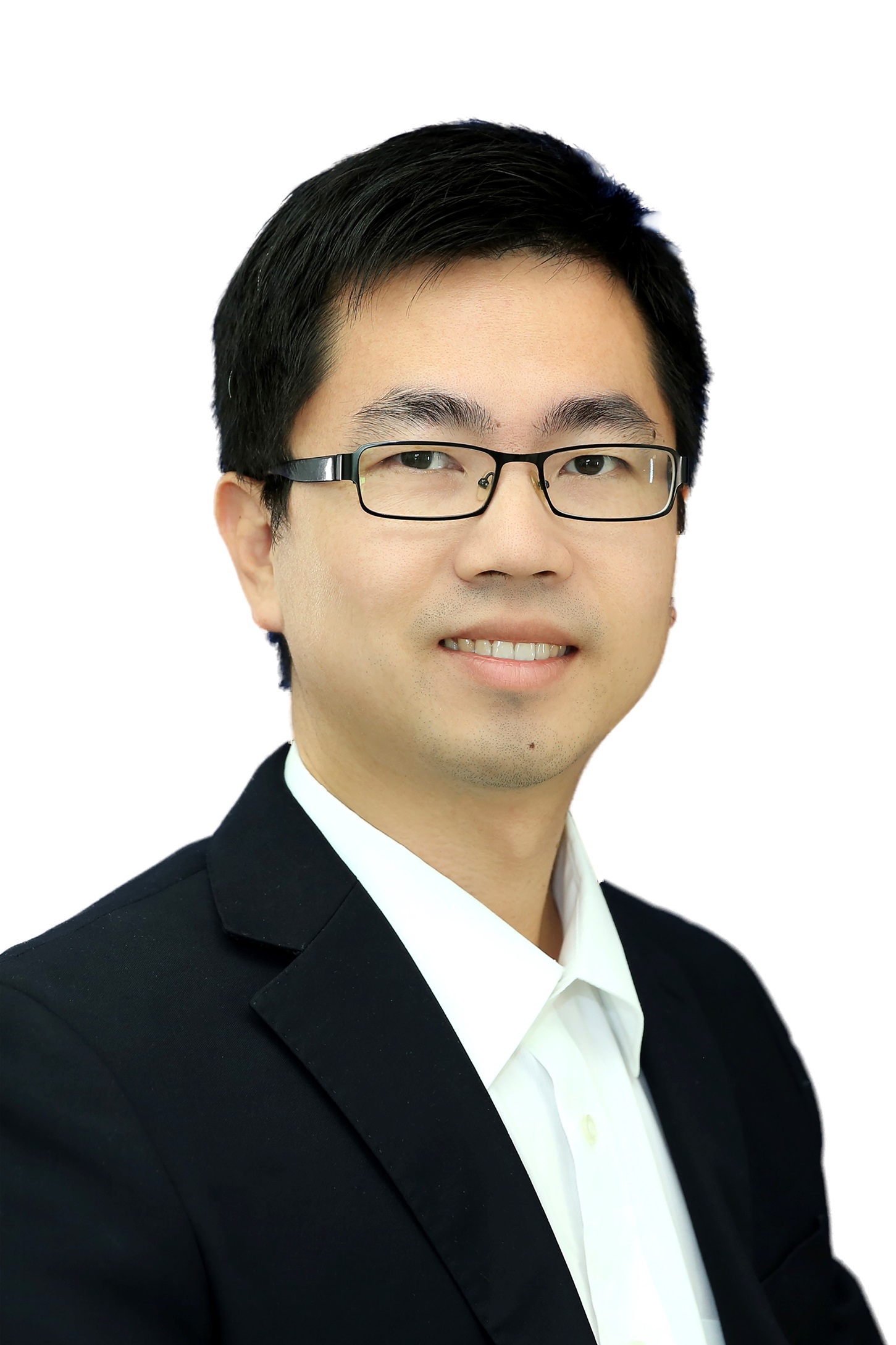}}]{Chao Dong} is a professor with Shenzhen Institute of Advanced Technology, Chinese Academy of Science (SIAT), and Shanghai AI Laboratory. In 2014, he first introduced deep learning method – SRCNN into the super-resolution field. This seminal work was chosen as one of the top ten “Most Popular Articles” of TPAMI in 2016. His team has won several championships in international challenges –NTIRE2018, PIRM2018, NTIRE2019, NTIRE2020 AIM2020 and NTIRE2022. He worked in SenseTime from 2016 to 2018, as the team leader of Super-Resolution Group. 
His current research interest focuses on low-level vision problems, such as image/video super-resolution, denoising, and enhancement.
\end{IEEEbiography}

\vfill

\end{document}